\documentclass[10pt,dvipsnames]{article}
\usepackage[utf8]{inputenc} 
\usepackage[T1]{fontenc}    
\usepackage{url}            
\usepackage[top=1in, bottom=1in, left=1in, right=1in]{geometry}
\usepackage{mathtools}
\usepackage{amsmath, amssymb, amsthm, bbm}
\usepackage{physics}
\usepackage{graphicx}
\usepackage{hyperref}       
\usepackage{caption}
\usepackage{tensor}
\usepackage[most]{tcolorbox}
\usepackage{empheq}
\usepackage{enumitem}
\usepackage{float}
\usepackage{graphicx}
\usepackage{booktabs}       
\usepackage{amsfonts}       
\usepackage{nicefrac}       
\usepackage{microtype}      
\usepackage{cleveref}
\usepackage{natbib}
\usepackage{tikz}
\usepackage{xfrac}
\usepackage{algpseudocode}
\usepackage{algorithm}
\usepackage{authblk}

\hypersetup{colorlinks, breaklinks=true, urlcolor=orange, linkcolor=blue, citecolor=blue}

\title{A theoretical perspective on mode collapse in variational inference 
}

\author[1,2]{Roman Soletskyi}
\author[1,2]{\;Marylou Gabrié}
\author[3]{\;Bruno Loureiro}

\def\relu{\text{relu}}

\affil[1]{\small 
CMAP, CNRS, École polytechnique, Institut Polytechnique de Paris, 91120 Palaiseau, France}
\affil[2]{\small 
Laboratoire de Physique de l’École normale supérieure, ENS, Université PSL,
CNRS, Sorbonne Université, Université de Paris F-75005 Paris, France}
\affil[3]{\small Département d’Informatique, École Normale Supérieure - PSL \& CNRS, France}

\date{\today}

\allowdisplaybreaks
\begin{document}

\maketitle
\begin{abstract}
While deep learning has expanded the possibilities for highly expressive variational families, the practical benefits of these tools for variational inference (VI) are often limited by the minimization of the traditional Kullback-Leibler objective, which can yield suboptimal solutions. A major challenge in this context is \emph{mode collapse}: the phenomenon where a model concentrates on a few modes of the target distribution during training, despite being statistically capable of expressing them all. In this work, we carry a theoretical investigation of mode collapse for the gradient flow on Gaussian mixture models. We identify the key low-dimensional statistics characterizing the flow, and derive a closed set of low-dimensional equations governing their evolution. Leveraging this compact description, we show that mode collapse is present even in statistically favorable scenarios, and identify two key mechanisms driving it: mean alignment and vanishing weight. Our theoretical findings are consistent with the implementation of VI using normalizing flows, a class of popular generative models, thereby offering practical insights. 
\end{abstract}

\section{Introduction}
\label{sec:intro}
Sampling is a fundamental task in science, arising across a variety of contexts such as machine learning \citep{teh2003energy}, Bayesian inference~\citep{mackay2002information} and statistical physics~\citep{Krauth2006}. Nevertheless, it can be a challenging problem, and understanding when a target distribution of interest can be efficiently sampled is largely an open question. Variational Inference (VI) addresses this problem by approximating the target distribution $p$ by minimizing the Kullback-Leibler (KL) divergence $D_{\textrm{KL}}(q || p)$ with respect to a conveniently chosen family  of tractable distributions $q\in\mathcal{Q}$. 

A major limitation of VI is the expressiveness of the variational family $\mathcal{Q}$. To ensure that $D_{\textrm{KL}}(q || p)$ is easy to evaluate, the family $\mathcal{Q}$ is restricted to distributions with a tractable normalization constant and/or easy to sample form. These requirements often imply turning to factorized distributions (leading to independent covariates) or Gaussian measures which are choices of $\mathcal{Q}$ that cannot be adequate in all interesting settings.    
In this context, normalizing flows~\citep{tabakDensityEstimationDual2010,papamakariosNormalizingFlowsProbabilistic2021a,kobyzevNormalizingFlowsIntroduction2021} and autoregressive models~\citep{larochelleNeuralAutoregressiveDistribution2011b,uriaNeuralAutoregressiveDistribution2016c} - both deep generative models with tractable normalized density and efficient sampling procedures - appear as promising variational families with an increased flexibility upon Gaussian or factorized measures. A series of work have already demonstrated the relevance of these deep learning approaches, both in the context of machine learning~\citep{rezendeVariationalInferenceNormalizing2015} and statistical physics~\citep{wu2019statvarinf,albergoFlowbasedGenerativeModels2019} problems. In the aforementioned examples, the KL divergence is typically minimized using a variation of gradient descent on the neural net parameters.  

Despite its popularity, it is well known that VI often encourages a mode-seeking behavior, capturing only one of the (typically) multiple target modes~\citep{regli2018alphabeta, jerfel_variational_2021,blessing2024elboslargescaleevaluationvariational}. This phenomenon, known as \emph{mode collapse}, is not resolved by using expressive variational families such as normalizing flows and autoregressive networks despite their ability to represent multi-modal distributions. In this context, solutions to mode collapse have been proposed, including annealing \citep{wu2019statvarinf}, adding a negative likelihood term of approximate samples from $p$ to the optimized  objective \citep{noeBoltzmannGeneratorsSampling2019} or dropping the VI objective altogether using an adaptive Monte Carlo framework instead \citep{gabrie_adaptive_2022}. Yet, to this date, the mechanisms behind mode collapse remain poorly understood. 

In this work, we take a first step towards closing this gap by investigating the dynamics of gradient-based optimization in the simplest setting of a bi-modal Gaussian mixture target distribution $p$. Our \textbf{main findings} are: 
\begin{itemize}
    \item First, we empirically show that mode collapse is present when learning this simple target model with different choices of variational families $q\in\mathcal{Q}$, such as normalizing flows and Gaussian mixtures. We identify two mechanisms, mean alignment and vanishing weights, that drive mode collapse in these experiments.
    \item Focusing on the simplest example of a Gaussian mixture models, we derive an exact description of the gradient flow dynamics for VI that allow us to study mode collapse in terms of the fixed points of a low-dimensional dynamical system for the relevant summary statistics. Leveraging this description, we mathematically investigate both the mean alignment and weight vanishing mechanisms identified in the experiments.
    \item Finally, we revisit mode collapse for normalizing flows from the perspective of our analytical findings, showing that they indeed share much of the phenomenology found for well-specified VI. We further investigate how initialization and specific parameterization of the variational distribution might help mitigating mode collapse. 
\end{itemize}

\paragraph{Related work ---} While mode collapse is a well known phenomenon, to the best of our knowledge, there exists very few systematic studies of where and why it arises. Recently, \cite{blessing2024elboslargescaleevaluationvariational} conducted a large scale empirical evaluation of VI approaches, reporting that mode collapse becomes harder and harder to avoid as the dimension increases, but without discussing mechanisms driving it.

Conversely, \cite{pmlr-v235-huix24a} derived theoretical guarantees on VI through the gradient flow on a mixture of Gaussian with fixed weights, a setting that is considered below, but did not focus on multi-modal target distributions and did not discuss the occurence of mode collapse.  

The loss landscape of Gaussian mixture models has been widely studied, both empirically and theoretically, in the context of maximum likelihood estimation (MLE) and expectation maximization \citep{Srebro2007, NIPS2016_3875115b, NIPS2016_792c7b5a, Fan2023, chen2024local}. We stress that this is a different problem from the one considered here. While the MLE assumes samples from the target probability are available such that all modes will be covered, in VI one cannot evaluate the target, and therefore resorts to the reverse KL learning objective estimated using exclusively samples from the variational model. In this context, some modes can be completely missed by the model.

Finally, the study of training dynamics through the lenses of a low-dimensional dynamical system for the correlation functions (or summary statistics) is a classical topic in the statistical physics of learning literature \citep{Kinouchi_1992, saad1995online, saad_1996}. More recently, it has been widely employed in the study of SGD for neural networks \citep{goldt_2019, refinetti21b, veiga2023phase, ben2022high, arous2023high, arnaboldi2023high, arnaboldi2023escaping, arnaboldi24a, collins2023hitting, patel2023rl, jain2024bias, mori2024optimal}. However, to our best knowledge this is the first work to use these ideas to the study of VI.

\section{Mode collapse in variational inference}
\label{sec:setting}
Consider a target distribution $p$ on $\mathbb{R}^{d}$ that we would like to sample from. Variational inference consists of finding an approximation of $p$ within a chosen family $q\in\mathcal{Q}$ of distributions on $\mathbb{R}^d$ by minimising the Kullback-Leibler (KL) divergence: 
\begin{equation}
\label{eq:def:kl}
    D_{\textrm{KL}}(q||p) = \mathbb{E}_q \log q(x) - \mathbb{E}_q \log p(x),
\end{equation}
where $\mathbb{E}_\nu$ stands for the expectation with respect to the distribution $\nu$. Note that for notational convenience we use the same notation for a distribution and its density with respect to the Lebesgue measure. The function $q \mapsto D_{\textrm{KL}}(q||p)$ is convex and, provided $p\in\mathcal{Q}$, it admits a unique minimum for $q=p$. However, even in this optimistic scenario optimising in the space of measures is computationally intractable. Therefore, the common practice consists of choosing a parametric family $\mathcal{Q} = \{q_\theta ; \theta \in \Theta\}$ and optimising $\mathcal{L}(\theta) = D_{\textrm{KL}}(q_\theta||p)$ over $\theta\in\Theta$. Our focus in this work is on the popular case where a Monte Carlo estimator of $\nabla \mathcal{L}(\theta)$ is available, and optimisation over $\theta\in\Theta$ is done by a descent-based method.

Note that the VI objective in \cref{eq:def:kl} consists of two terms. The first is the entropy, and favours distributions with lighter tails. On the other hand, the second term vanishes if $q_{\theta} = 0$ where $p > 0$. Together, this results in a mode-seeking behavior~\citep{regli2018alphabeta,miller2017variational}: for multi-modal $p$, the optimal $q$ might approximate only a subset of the modes and, as a result, not cover the full $p$ distribution.\footnote{Note that, on the contrary, minimising the direct KL divergence $ D_{\textrm{KL}}(p||q)\neq  D_{\textrm{KL}}(q||p)$ results in mass-covering behavior. Indeed, if $q$ does not cover some mode and vanishes where $p > 0$, the second term gives an infinite penalty. However, estimating the direct KL divergence requires sampling from $p$ which is, by construction in VI, computationally hard in practice.}  As a consequence, VI is prompt to mode collapse. Next section illustrates this discussion on a few preliminary experiments, showing that mode collapse is present even when $p\in\mathcal{\mathcal{Q}}$.

\begin{figure}[t]
    \centering
    \includegraphics[width=0.6\linewidth]{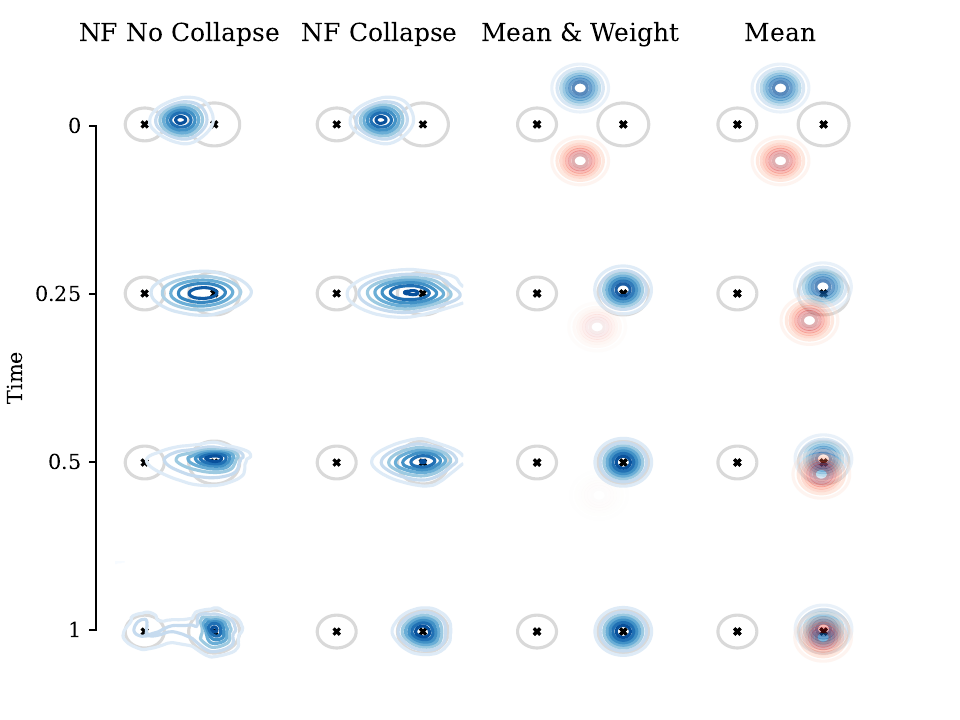}
    \caption{Evolution of gradient descent dynamics from initialization $T=0$ to convergence $T=1$ for VI on a 2-Gaussian mixture target distribution $p(x)$ as in \cref{eq:pstar2d}
    with $||\mu_\star||_2 = 2.5$ for the first column, $||\mu_\star||_2 = 3.1$ for the remaining ones, and $w_\star = \sfrac{1}{3}$ for all the columns. The mode position is marked by a black cross, and the 9th decile by gray lines. (\textbf{First and second}) $q_{\theta}$ depicted in blue is parameterized by a normalizing flow, see \cref{sec:app:normalizing}. (\textbf{Third}) $q_{\theta}(x)=w_{1}\mathcal{N}(\mu_{1},I_{2})+w_{2}\mathcal{N}(\mu_{2},I_{2})$ where we optimize over both means and weights $\theta=(w_{1},w_{2},\mu_{1},\mu_{2})$. (\textbf{Fourth}) $q_{\theta}(x)=w_\star\mathcal{N}(\mu_{1},I_{2})+(1-w_\star)\mathcal{N}(\mu_{2},I_{2})$ where we optimize only over the means $\theta=(\mu_{1},\mu_{2})$. In the last two columns the density the different components of $q_\theta$ are depicted in different colors and their opacity is proportional to their weight.}
    \label{fig:mode-collapse}
\end{figure}

\paragraph{Preliminary experiments ---}
Our starting point is the minimal example of a multi-modal target distribution $p$: a mixture of two isotropic Gaussians in dimension $d=2$: 
\begin{equation}
    \label{eq:pstar2d}
    p = w_\star\mathcal{N}(\mu_\star,I_{2})+(1-w_\star)\mathcal{N}(-\mu_\star, I_{2}),
\end{equation}
where $\mathcal{N}(\cdot,\cdot)$ stands for the Gaussian distribution, $\mu_\star$ is a mean vector chosen such that the modes are well-separated and 
$w_\star\in (0, \sfrac{1}{2})$
such that the mixture is unbalanced. \Cref{fig:mode-collapse} reports the evolution of the gradient descent dynamics on the VI objective for different classes of $q_{\theta}\in\mathcal{Q}$.
While the normalizing flow (NF) can represent the bi-modal target (first column), increasing the distance between target clouds can lead to mode collapse (second column). 
We simplify further the experiment by considering as $q_\theta$ a mixture of two Gaussians with the components means and weights as tunable parameters, leaving the covariances equal to identity. Here, we observe that mode collapse is typically the result of one of the components weights going to zero following the alignement of the other component with one of the target's components (third column). Lastly, we fix the balance of the weights of the variational model to be equal to the balance in the target and run the gradient descent solely on the means. Mode collapse can also happen in this setting with the two means flowing to the same target component (last column).



These preliminary experiments show that mode collapse is present even in the simplest multi-modal VI tasks, and unveils two possible mechanisms behind it: weight vanishing and mean alignment. Normalizing flows do not have a well defined bi-modality as the bi-model Gaussian model but their behavior seems to be closer to the former case: two clouds start to form but in the end only one prevails. 
Our goal in the following is to develop a mathematical understanding of these two mechanisms for mode collapse in the simpler Gaussian mixture model.
\section{Variational inference with Gaussian mixtures}
\label{sec:dynamics}
In this section, we consider the minimal model for mode collapse discussed in \cref{sec:setting}, a mixture of two isotropic Gaussian clouds in $\mathbb{R}^{d}$:
\begin{align}
    & p(x) = w_{\star}\mathcal{N}(x|\mu_{\star}, I_{d}) + (1-w_{\star}) \mathcal{N}(x|-\mu_{\star}, I_{d}), \notag\\
    \label{eq:setting:q}
    & q_{\theta}(x) = w_{1} \mathcal{N}(x|\mu_{1}, I_{d})+ w_{2} \mathcal{N}(x|\mu_{2}, I_{d}),
\end{align}
with $w_{\star}\in(0,1)$, $w_{1}+w_{2}=1$ and without loss of generality we assumed the target 
means to be symmetric around the origin.
Note that although $p,q_{\theta}$ belong to the same parametric family, we denote the variational distribution $q_{\theta}$ to stress that $\theta$ = $\{w_{1},w_{2},\mu_{1},\mu_{2}\}\in\mathbb{R}^{2(d+1)}$ are the trainable parameters. Note that the discussion in this section is readily generalized to a variational family of Gaussian mixtures with more modes, but to simplify the exposition we focus here in the bi-modal case, referring the interested reader to Appendix \ref{sec:app:derivation} for the general case. We denote $R^{2}=||\mu_{\star}||^{2}_{2}$ and define for convenience the ratio of target weights $\gamma \coloneqq w_{\star} / (1-w_{\star})$. 

As discussed in \cref{sec:setting}, variational inference consists of minimising the following objective:
\begin{equation}
    \label{eq:setting:loss}
    \underset{\theta}{\min}\left[\mathcal{L}(\theta) \coloneqq D_{\textrm{KL}}(q_\theta||p)\right].
\end{equation}
Our main goal in this section is to investigate the gradient flow dynamics:
\begin{equation}
    \label{eq:gf}
    \dot{\theta} = -\nabla_\theta\mathcal{L}.
\end{equation}
Note that in practice one considers instead gradient descent at small learning rate $\eta>0$ and an empirical approximation of the loss function \cref{eq:setting:loss} over a large finite batch $B$ from $q_{\theta}$, which is indeed what is considered in the numerical experiments. 

\subsection{Fixed weight dynamics}
\label{sec:fixedweight}
The empirical results from \cref{sec:setting} suggest that mode collapse can be driven either by vanishing weights $w_{1}, w_{2}$ or by mean alignment. In order to study these separately, we start our investigation by considering a fixed weight scenario where $w_{1},w_{2}$ are fixed and only the means are trained. To simplify the analysis further, we fix the norm of the means $\mu_{1},\mu_{2}\in\mathbb{S}^{d-1}(R)$ and consider a spherical flow:
\begin{equation}
    \label{eq:dynamics:sphere-grad}
    \dot{\mu}_{c} = \nabla_{\mu_{c}}^{\mathbb{S}}\mathcal{L} = \left(I_{d} - \frac{\mu_{c}\mu_{c}^{\top}}{R^2}\right)\nabla_{\mu_{c}}\mathcal{L}, \quad c=1,2 \,.
\end{equation}
Although the analysis can be equally carried for standard gradient flow, we have found that the evolution of the norm does not play a major role in the presence (or lack of thereof) of mode collapse, and we refer the interested reader to Appendix \ref{sec:app:derivation} for details on the Euclidean flow. The key idea in our analysis is to notice that mode collapse can be entirely characterized by the evolution of the following three correlations:
\begin{align}
    m_{1}=\frac{\mu_{1}^{\top}\mu_{\star}}{R^{2}}, && m_{2}=\frac{\mu_{2}^{\top}\mu_{\star}}{R^{2}}, && s = \frac{\mu_{1}^{\top}\mu_{2}}{R^{2}}
\end{align}
This allows us to reduce the high-dimensional evolution of means $\mu_{1},\mu_{2}\in\mathbb{S}^{d-1}(R)$ in \cref{eq:dynamics:sphere-grad} to a low-dimensional evolution in terms of sufficient statistics $m_{1},m_{2},s\in[-1,1]$:
\begin{align}
    \label{eq:dynamics:flow}
    \dot{m}_{1} &= -\left[(m_2 - m_1 s)f(s) + w_{1}(1 - m_1^2)g(m_1)\right] \notag\\ 
    \dot{m}_{2} &= -\left[(m_1 - m_2 s)f(s) + w_{2}(1 - m_2^2)g(m_2)\right] \notag\\
     \dot{s} &= -\left[2(1 - s^2)f(s) + w_{1}(m_2 - m_1 s)g(m_1) +  w_{2}(m_1 - m_2 s)g(m_2)\right]
\end{align}
where the auxiliary functions $f,g:[-1,1]\to\mathbb{R}$ are given by:
\begin{align}
    \label{eq:dynamics:f}
    f(s) &= \mathbb{E}_{z}\left[w_{1}\sigma\left(R^2(s - 1) + z R\sqrt{2(1 - s)}- \log\frac{w_{1}}{w_{2}}\right)^2 + w_{2}\sigma\left(R^2(s - 1) + z R\sqrt{2(1 - s)} + \log\frac{w_{1}}{w_{2}}\right)^2\right] \notag\\
    g(m) &= 1 - 2\mathbb{E}_{z}\left[\sigma\left(2R^2m + 2Rz + \log \frac{w_{\star}}{1 - w_{\star}}\right)\right]
\end{align}
with $z\sim\mathcal{N}(0,1)$ and $\sigma(t)\coloneqq(1+e^{-t})^{-1}$ denote the sigmoid function. Note that $g$ is a monotonic function.
The detailed derivation can be found in Appendix~\ref{appendix:stat-evolution}.

\subsubsection{Fixed points of the dynamics}

Since \cref{eq:dynamics:sphere-grad} is a descent algorithm, it converges to zero gradient points of $\mathcal{L}$ (provided they exist). The low-dimensional equations \cref{eq:dynamics:flow} provide a convenient way of classifying these fixed points and detecting mode collapse. In the following, we will say the dynamics has mode collapsed if it converges to a fixed point $(m_{1}^{*}, m_{2}^{*}, s^{*})$ such that $s^*>0$.

Consider the case where the weights of the variational distribution are fixed to their optimal value $w_{1}=(1-w_{2})=w_{\star}$. This is a statistically favorable scenario where the variational model is well-specified, initialized with the correct target weights and 
norm $R$. Yet, as we will see now, mode collapse is still present.  More precisely, in Appendix~\ref{appendix:fixpt} we prove that \cref{eq:dynamics:flow} has three types of fixed points:
\begin{enumerate}[label=(\roman*), leftmargin=*]
    \item \textbf{Global minimun and flipped fixed point:}  The global minima of the loss is given by $\mu_{1}=-\mu_{2}=\mu_{\star}$ which corresponds to $(m^*_{1},m^*_{2},s^*)=(1, -1, -1)$. Flipping the role of $\mu_{1}$ and $\mu_{2}$ yields another fixed point of \cref{eq:dynamics:flow}, $(m^*_{1},m^*_{2},s^*)=(-1, 1, -1)$, which is also a global minima of the loss in the symmetric case $w_\star=0.5$.
    \item \textbf{Perfect alignment mode collapse:} Corresponds to $\mu_{1}=\mu_{2}$ and hence to $s^*=1$. This is a fixed point of \cref{eq:dynamics:flow} provided $m^*_{1}=m^*_{2}=m$ and $m=\pm 1$ or $g(m)=0$. Since $g$ is monotonic in $[-1,1]$, it has a unique root. Hence, there are 3 perfect alignment fixed points. 

    \item \textbf{Other fixed points:} Other fix points exists with $m^*_1 \neq m^*_2$ and $s^*\neq1$, but a closed-form expression is not available.
    Nevertheless, they can be easily studied numerically by computing zeroes of \cref{eq:dynamics:flow} over $(m_{1},m_{2},s)\in(-1,1)^{3}$. In particular, mean alignment mode collapse corresponds to fixed points where $s^*\in(0,1)$.
\end{enumerate}
To determine whether a fix point $(m_{1}^{*},m_{2}^{*},s^{*})\in[-1,1]^{3}$ is an attractor or repeller of the flow, we must study the spectrum of the Hessian at the fixed point. In particular, if the smallest eigenvalue is positive, the fixed point is attractive/stable, and corresponds to a minimum of the loss. 

Remarkably, the eigenvalues of the Hessian can be explicitly computed for the scenarios (i), (ii) discussed above - see Appendix~\ref{appendix:fixpt}. In these cases, we can show that the only stable fixed points are the global minimum and above a certain $R$ and flipped fixed points, both corresponding to a non-collapsed outcome with $s^*=-1$. The perfect alignment mode collapse fixed points $s^{*}=1$ instead are always unstable. 

Searching for other fixed points with $s^{*}\neq 1$, one finds numerically that below a certain critical threshold ${R<R_{c}}$, (i), (ii) are the only fixed points of the flow \cref{eq:dynamics:flow}. This means that for small radius, the mode collapse fixed point is always unstable, and hence not reached dynamically. 

On the other hand, above the critical threshold $R>R_{c}$ an additional fixed point with $s^{*}\in(0,1)$ appears. One can check numerically that this mean alignment mode collapse fixed point is stable. Interestingly, despite being stable this fixed point is not universally attracting in the $(m_1, m_2, s)$-plane, meaning that whether the flow in \cref{eq:dynamics:flow} converges to it or to the global or flipped minimum, thereby avoiding mode collapse, depends on the initialization of the means $\mu_{1},\mu_{2}$. To lead further the discussion, we consider next the high-dimensional scenario where the initialization of the summary statistics likely lies close to the origin.

\subsubsection{High-dimensional scenario}
\label{subsec:high-dim}
Typically, the VI practitioner has no information on the ground truth means $\mu_{\star}$, and therefore initializes the algorithm independently at random $\mu_{1},\mu_{2}\sim{\rm Uni}(\mathbb{S}^{d-1}(R)$). Therefore, with high-probability one has $m_{1},m_{2},s = O(\sfrac{1}{\sqrt{d}})$ at initialization. 

\begin{figure}[t]
    \centering
    \includegraphics[width=0.6\linewidth]{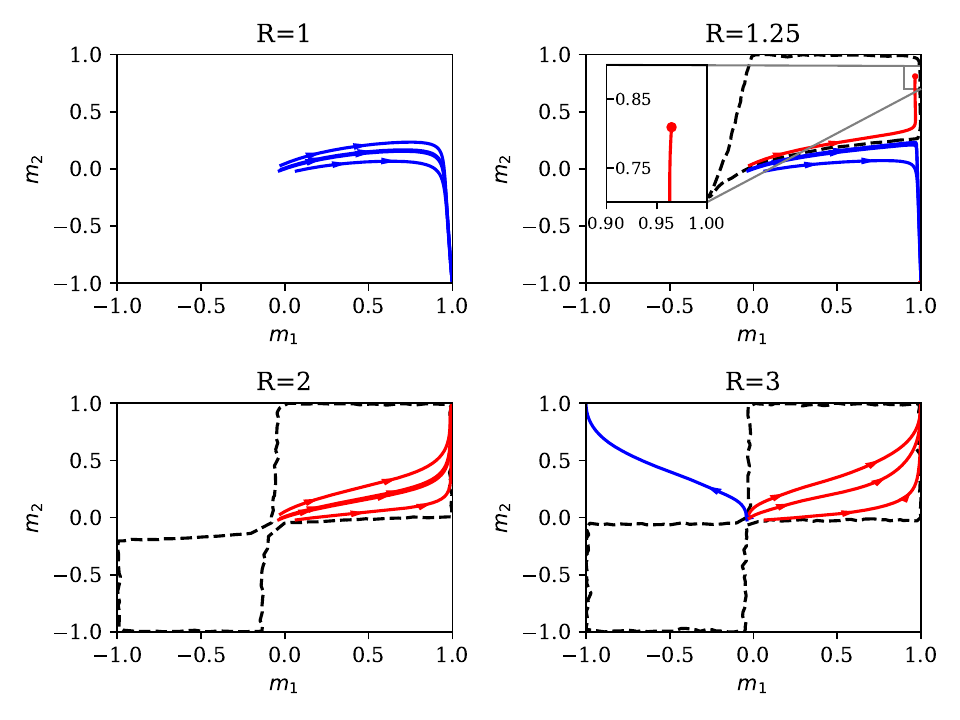}
    \caption{Basin of attraction of the mean alignment fixed points $m_{1}=m_{2}=\pm 1$ on the $s=0$ cross section of phase space $(m_{1},m_{2},s)\in[-1,1]^{3}$ for $R\in\{1, 1.25, 2, 3\}$, $d = 1000$, $w_{\star} =\sfrac{2}{3}$, and $\eta=0.05$. The dashed black line denote the boundary of the basin of attraction of the mode collapse fixed points. Solid lines denote individual flow trajectories with different random initialization $\mu_{1},\mu_{2}\sim{\rm Unif}(\mathbb{S}^{d-1}(R))$. Mode collapsed trajectories are red, avoiding it are blue. }
    \label{fig:basin}
\end{figure}

In \cref{fig:basin}, we illustrate the typical trajectories of the summary statistics in the plane $(m_1, m_2)$ for uniformly random initialization of the means on the $d$-dimensional sphere and increasing values of the radius $R$. Consistently with the above stability analysis, below $R_c$ the dynamics converge to the global minimum. For $R>R_c$, the outcome depends on initialization even in this high-dimensional scenario.

For $R>R_c$, we investigate numerically the basin of attraction of the mode collapse stable fixed points, that is with $s^*\in (0,1)$ (scenario (iii)). The dashed black line represents the boundary of their basin of attraction at a $s=0$ slice of the phase space $(m_{1},m_{2},s)$: under the flow \cref{eq:dynamics:sphere-grad}, any initial condition $(m^{0}_{1},m^{0}_{2}, 0)$ inside this region necessarily flows to the mode collapse mean alignment fixed point. Interestingly,  
the boundaries of the basin attraction lies very close to the origin for some directions meaning that fluctuations can lead to initialization outside the basin, even in high-dimension. 
Whether a trajectory will converge to the global or flipped minimun or mode collapse therefore crucially depends on the initialization even at large $d$. Moreover, \cref{fig:basin} suggests that for large $R$, the basin of attraction of mode collapse consists of the symmetric quadrants $m_{1}m_{2}>0$.
    This behavior can be mathematically understood by noting that $f(s)$ in \cref{eq:dynamics:f} is $O(e^{-2(1-s)R^2})$ when $s < 1$. Hence, assuming $s$ remains smaller than $1$, 
    $f(s)$ is exponentially small along the flow trajectory, and the evolution of $(m_{1},m_{2})$ is approximately autonomous:
\begin{equation}
    \dot{m}_{1,2} \approx -w_{1,2}(1 - m_{1,2}^2)g(m_{1,2}) \, .
\end{equation}
Since $g$ is a decreasing function, once $|m|$ starts to increase, it monotonically continues along the entire trajectory. Therefore, for $R$ large enough, the final value of $m$ is completely determined by the sign of $g(m)$ at initialization. Note that the sign of $g(m)$ only changes in a $O(1/R^2)$ neighborhood of $m=0$. Since under random initialization $m^{0}=O(\sfrac{1}{\sqrt{d}})$ with high-probability, when $1/R^2\lesssim 1/\sqrt{d}\Rightarrow d\lesssim R^4$, there is a finite probability that at initialization $g(m_{1,2})$ have different signs and mode collapse is avoided. 

Therefore, in this context a simple strategy to avoid mean alignment mode collapse for systems with moderate dimensions $d$ is to restart the training with another random seed. 

\subsection{Training the weights}
\label{subsec:weight-evolution}
In the previous section, we discussed how mode collapse can dynamically arise even in the statistically favorable scenario where the weights are fixed at their target optimal values, driven by an attractive mean alignment mechanism. In this section, we investigate how training the variational model weights $(w_{1},w_{2})$ impacts this discussion. 

First, note that the naive flow in \cref{eq:gf} does not preserve the normalization of the weights $w_{1}+w_{2}=1$. Different ways of incorporating this constraint in the flow will lead to different optimization algorithms, which might be more or less prompt to mode collapse. In this manuscript, we focus on the following reparametrization of the weights:
\begin{align}
    \label{eq:weight:param}
    w_{1} = \frac{v_1}{v_1 + v_2}, && w_{2} = \frac{v_2}{v_1 + v_2};
\end{align}
for positive $v_{1},v_{2}\in \mathbb{R}_{+}$. In Appendix \ref{appendix:weight-evolution}, we show that this parametrization for the weights lead to the following normalization preserving flow on the weights:
\begin{align}
\label{eq:weight:flow}
\dot{w}_1 = -(w_{1}^2 + w_{2}^2)\left(\frac{\partial\mathcal{L}}{\partial w_{1}} - \frac{\partial\mathcal{L}}{\partial w_{2}}\right), && \dot{w_{2}} = -\dot{w}_1
\end{align}
where 
\begin{align}
    \frac{\partial\mathcal{L}}{\partial w_{1}} &= \mathbb{E}\left[\sigma\left(R^2(1 - s) + z R\sqrt{2(1 - s)}+ \log \frac{w_{1}}{w_{2}}\right)\right]+\frac{w_{2}}{w_{1}}\mathbb{E}\left[\sigma\left(-R^2(1 - s)+ z R\sqrt{2(1 - s)} + \log \frac{w_{1}}{w_{2}}\right) \right]\notag\\ 
    &\quad- \mathbb{E}\left[\log\sigma\left(R^2(1 - s) + z R\sqrt{2(1 - s)}+ \log\frac{w_{1}}{w_{2}}\right)\right] +\mathbb{E}\left[\log\sigma\left(2R^2m_1 + 2R z + \log\frac{w_{\star}}{1 - w_{\star}}\right)\right]\notag\\
    &\quad+ \log \frac{w_{1}}{w_{\star}} + R^2(1 - m_1)
\end{align}
where the expectations are over $z\sim\mathcal{N}(0,1)$. A similar expression for the derivative with respect to $w_{2}$ is obtained by noting that $w_{2}=1-w_{1}$. Therefore, the dependence on the means $\mu_{1},\mu_{2}$ in the flow on the weights in \cref{eq:weight:flow} is also entirely captured by the correlation functions $(m_{1},m_{2},s)$. 

\begin{figure}[t]
    \centering
    \includegraphics[width=\linewidth]{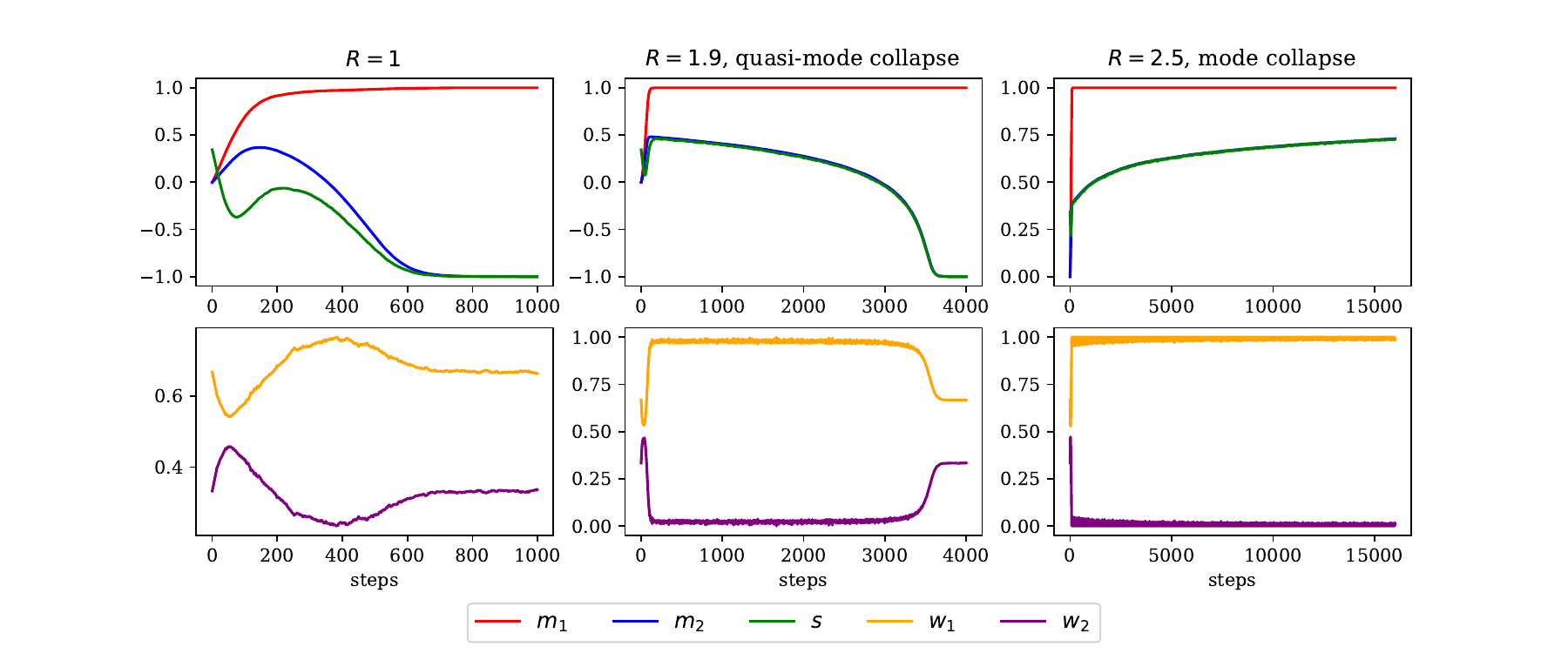}
    \caption{Numerical evolution of sufficient statistics and weights for different radii $R\in\{1,1.9, 2.5\}$. Quasi-mode collapse happens in the middle. Learning rate $\eta=0.05$, batch size $B=1000$, $d=10$ and $w_{\star}=\sfrac{2}{3}$.}
    \label{fig:quasi}
\end{figure}

In the Appendix \ref{appendix:proj-grad} we also discuss an alternative descent algorithm that projects the weights in the constraint $w_{1}+w_{2}=1$ at every step. In particular, we observe that this projective procedure is more prompt to weight vanishing mode collapse than the reparametrization in \cref{eq:weight:flow}, highlighting the importance of the implicit algorithmic bias to mode collapse in VI.
\paragraph{Quasi-mode collapse ---}
As we saw in the fixed weight dynamics, the existence of stable mode collapse fixed points depend on the distance between target means $R$. We start our discussion by numerically observing that a similar phenomenology arises when training the weights. \Cref{fig:quasi} shows the evolution of the correlation functions $(m_{1},m_{2},s)$ (top) and weights $(w_{1},w_{2})$ under the flow \cref{eq:weight:flow}, for increasing radii $R$. Again, we numerically observe the existence of a critical separation $R_{c}$ above which $s\in(0,1]$ is a fixed point of the flow, signaling the presence of mode collapse.
Note, however, that the mechanism leading to mode collapse is very different and shows its signature even for $R<R_{c}$:
early in the dynamics both the means $\mu_{1,2}$ positively correlates with the same target mode. Once, one of them reaches full alignment with this target modes, the opposite weight starts to decrease, and the evolution of the corresponding mean slows down. This can be understood from the fact that the loss terms in the evolution of the means are proportional to the weight, see \cref{eq:dynamics:f}. For small enough radius $R$ (\cref{fig:quasi} left), the small weight eventually recovers, and the dynamics quickly converge to the global minima. However, as the radius is increased (\cref{fig:quasi} middle), the slow dynamics becomes longer, meaning the system is trapped in an increasingly longer transient state with vanishing weight. We refer to this phenomena as \emph{quasi-mode collapse}. Finally, above $R>R_{c}$ this weight-vanishing mode collapse becomes attractive (\cref{fig:quasi} right).

We now investigate this empirical observation through the lenses of the summary statistics. First, note that $\mu_{1}$ becomes quickly aligned with $\mu_{\star}$, then $m_1\approx 1$ and from the problem's geometry $m_2\approx s$. Therefore, the flow of sufficient statistics and weights becomes effectively reduced to the flow of two variables $(s, w)$ where we have defined $w$ as smaller weight $w_{2}$. Inserting $m_1 = 1, m_2 = s$ into~\cref{eq:dynamics:flow} and expanding in $w\ll 1$ in~\cref{eq:weight:flow} we obtain:
\begin{align}
    \label{eq:quasi:flow-w}
    & \dot{w} = \frac{\partial\mathcal{L}}{\partial w_{1}} - \frac{\partial\mathcal{L}}{\partial w_{2}} \\
    \label{eq:quasi:flow-s}
    & \dot{s} = -(1-s^2)(2f(s)+w g(s))
\end{align}
Again, from our empirical observation, the evolution of $s$ is much slower than the evolution of $w$. Thus, we can treat $w$ as fast mode that thermalizes quickly and $s$ as a slow mode. Then, for most of the trajectory  $w$ can be approximated by its equilibrium value $w_{\textrm{eq}}(s)$ which slowly changes with $s$. We can find $w_{\textrm{eq}}$ by solving the stationary flow equation $\dot{w} = 0$ at large $R$. Consider the following \emph{ansatz} for the expansion $w$ at large $R$:
\begin{equation}
    \log w_{\textrm{eq}} = aR^2 + bR + \ldots
\end{equation}
The coefficients $a,b$ can be determined by inserting this ansatz in the flow equation and approximating $\log\sigma(x) = -\relu(-x)$ at large $|x|$. As a result, we find that when $s < 0$, up to a constant:
\begin{equation}
    \log w_{\textrm{eq}} = -R^2(1 + s).
\end{equation}
This allow us to compute $f(s)$ at large $R$. Using Laplace method, one finds $f(s)\propto e^{-R^2/(1-s)}$. Therefore, when $s > 0$, $g(s)\approx 1$ at sufficiently large $R$. Inserting this in \cref{eq:quasi:flow-s}, we obtain the following approximate dynamics for $s$:
\begin{align}
    \dot{s} &= -e^{-R^2(1-s)}
\end{align}
which admits a closed-form solution $s = -1 + \sfrac{1}{R^2}\log(e^{R^2} - t)$. This give us an estimation of the time scale over which the flow stays in a quasi-mode collapse state before recovering $T_{\textrm{quasi}}=O(e^{R^2})$.

Note that the simplification above, concerning the separation between fast $w$ and slow $s$ modes, holds for the flow at large radius. However, when comparing with numerical experiments we have observed that the fluctuations induced by a finite learning rate $\eta>0$ in the discretization of the flow can become relevant to the picture. Analyzing this interplay is beyond the scope of the present work, and is an interesting avenue for future work.
 
\section{Numerical experiments}
\label{sec:numerics}
\Cref{sec:dynamics} provided some understanding of the mechanisms behind mode collapse when performing VI using gradient descent. In this last section, we compare the theoretical findings obtained on Gaussian mixtures with the practical case of using NFs for VI on a multi-modal target distribution. 

\paragraph{Experimental setup ---}

The target distribution $p$ is kept to be equal to the bimodal Gaussian mixture considered in \cref{sec:dynamics} in dimension $d$. Without loss of generality, we consider the target means to be $\mu_\star=(R, 0, \ldots, 0)$ and $-\mu_\star$. 

The variational model considered here is a NF of RealNVP-type~\citep{dinh2017density}: $q_\theta$ is the pushforward of a base distribution $p_z$ by a diffeomorphish $f_\theta$, carrying the parameters $\theta$ of the VI. The probability density is explicitly given by the change of variable formula
\begin{equation}
    q_\theta(x) 
    = p_z(f^{-1}_\theta(x))\left\lvert\nabla_x f^{-1}_\theta\right\rvert
\end{equation}
where the building blocks of the RealNVP map $f_\theta$, called coupling layers, are designed to allow for easy computation of Jacobians. The reverse KL is minimized by stochastic gradient descent on $\theta$ leveraging an unbiased Monte Carlo estimator of $\nabla \mathcal{L}(\theta)$ using a batch of size $B$ of samples from the base distribution $p_z$. Details on the architecture and loss computation are deferred to \cref{appendix:numerics}. 

In the experiments below, $f_\theta$ is initialized close to the identify, such that at the beginning of the learning $q_\theta \approx p_z$. Thus we can investigate different strategy of initialization by considering different base distributions: the \emph{centered} case where $p_z = \mathcal{N}(0,I_d)$; the \emph{shifted} case  $p_z = \mathcal{N}(\mu,I_d)$ with $\mu$ a random vector with norm $R$\footnote{This setup is closer to the Gaussian mixture theory because at initialization the center of the Gaussian cloud approximately lies at the sphere of radius $R$.} and finally the \emph{multi-modal prior} case
\begin{equation}
    p_z = w_\star\mathcal{N}(\mu, I) + (1 - w_\star)\mathcal{N}(-\mu, I)
\end{equation}
where $\mu$ is again a random vector with norm $R$. The latter case closely resembling the fixed weight setting studied in \cref{sec:fixedweight}, at least at initialization.

To describe the mode collapse phenomenon for NFs, we partition $\mathbb{R}^d$ into two half-spaces $H_{\pm}$ according to the sign of the first coordinate and define the statistics
\begin{align}
    \label{exp:nf:weights}
    & w_{\pm} = \mathbb{E}_{q_\theta}\left[\mathbf{1}_{x\in H_{\pm}}\right] \, ,\\
    & \mu_{\pm} = \frac{1}{w_{\pm}}\mathbb{E}_{q_\theta}\left[x \mathbf{1}_{x\in H_{\pm}} \right] \, ,
\end{align}
along with the projections $m_{\pm} = \mu_{\pm}^{\top}\mu_{*}$ and $s=\mu_{+}^{\top}\mu_{-}$.
In the experiments below, we consider that mode collapse has occurred when at the end of the optimization $s > 0$ or one of $w_{\pm} < 0.01$. 

As a reference, we also report here on the Gaussian mixture setup of \cref{subsec:weight-evolution} initializing means as random vectors $\mu_{1,2}\sim{\rm Unif}(\mathbb{S}^{d-1}(R))$ and weights as $w_{1} = w_\star$ and $w_{2} = 1 - w_\star$. As above, we consider that mode collapse has occurred when at the end of the optimization $s > 0$ or one of $w_{1,2} < 0.01$. 

\begin{figure}[t]
    \centering
    \includegraphics[width=0.6\linewidth]{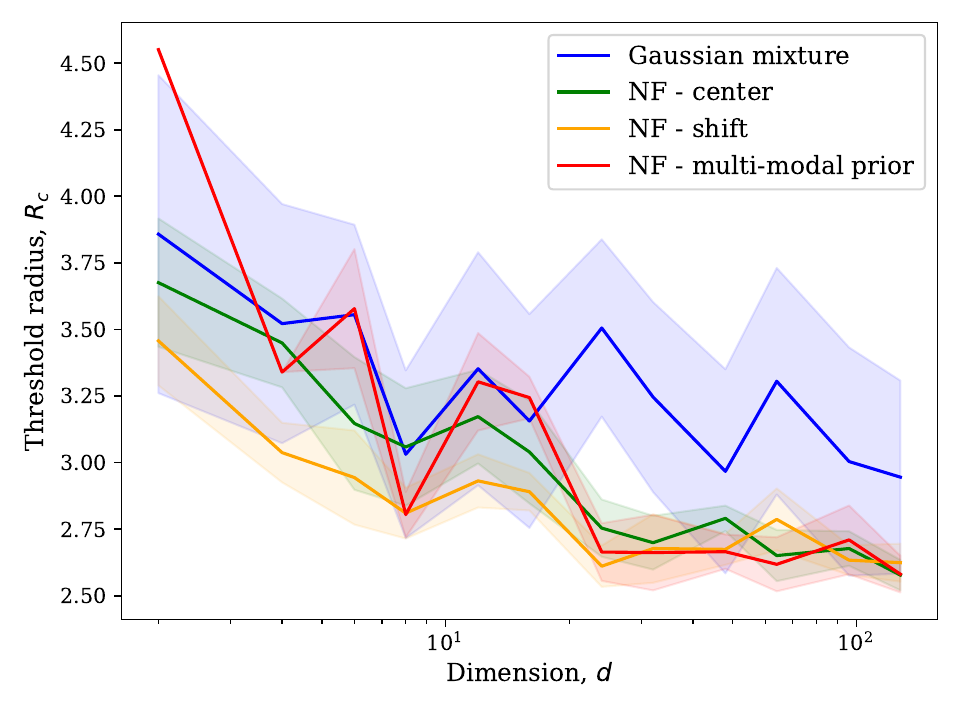}
    \caption{Dependence of threshold radius $R_c$ on dimension $d$ for different setups
    }
    \label{fig:threshold}
\end{figure}

\paragraph{Results ---} We find that the observation made on Gaussian mixtures that mode collapse occurs when the radius $R$ is larger than a threshold $R_c$ is also generally true for NFs. To further quantify this agreement, we perform a binary search on the radius to locate the threshold $R_c$ as a function of the dimension $d$. The binary search is stopped when a tolerance of 0.01 is reached. Since mode collapse can depend on the random initialization, each experiment is repeated with 10 different seeds. The resulting thresholds for NFs with the 3 different priors and the mixture of Gaussian variational family
are plotted on \cref{fig:threshold}. 

Interestingly, for the multi-modal prior, we observe that for all tried dimensions $d$, up to $d = 128$, there are seeds that avoid mode collapse at all $R$. For this settings, the radius plotted on \cref{subsec:weight-evolution} only includes seeds that do experience mode collapse. Therefore, the multi-modal prior setup can be beneficial to avoid mode collapse at the cost of having to restart the experiment with a different seed. This observation is consistent with the discussion on Gaussian mixtures with fixed weights of \cref{subsec:high-dim}, such that we expect it to hold at least for moderate dimensions $d$. 

For the two other NF settings, the threshold $R_c$ are very similar, especially in large dimension. The Gaussian mixture model enjoys a slightly larger threshold but the qualitative picture is consistent with the NFs: $R_c$ has a finite value that decreases as $d\rightarrow\infty$. This is another indicator of the similarity between the mechanisms behind mode collapse for the Gaussian mixture variational family and the NFs.

\section{Conclusion}
In this work, we have presented a theoretical investigation of mode collapse in variational inference, focusing on the gradient flow dynamics in Gaussian mixture models. Our results build on an exact characterization of the flow in terms of low-dimensional summary statistics. This reduced description maps the study of mode collapse to the study of the fixed points of this low-dimensional dynamical system. 

Leveraging this description, we identified two principal mechanisms behind mode collapse: mean alignment and vanishing weights. These mechanisms were shown to drive the collapse even in scenarios where the variational family is statistically well-specified, corroborating the observation that mode collapse is not a by-product of insufficient model expressivity, but a consequence of the optimization procedure. 

Additionally, the separation between modes was shown to play a crucial role in determining whether or not mode collapse occurs. This observation was found to be consistent when considering the more powerful variational family of normalizing flows.

The observation that expressivity is not a driving factor of mode collapse suggests that the investigation of mathematically tractable multi-modal variational families, such as the Gaussian mixture model studied here, offers a promising road map towards understanding, and hence mitigating, mode collapse in more complex tasks. The qualitative agreement we observed between theoretical predictions and numerical experiments on more sophisticated models further confirm the relevance of this approach. Our results lay the groundwork in this direction.
\subsubsection*{Acknowledgements} 
We would like to thank Louis Grenioux and Zhou Fan for stimulating discussions. BL acknowledges funding from the \textit{Choose France - CNRS AI Rising Talents} program. MG and RS acknowledge funding from Hi! Paris.

\bibliographystyle{plainnat}
\bibliography{bibliography}

\newpage
\appendix
\section*{\LARGE Appendix}
\section{Derivation of summary statistics evolution}
\label{sec:app:derivation}
\subsection{Recap of the setting}
For the reader convenience, we start by recalling the setting. We are interested in studying the following \emph{variational inference} (VI) objective:
\begin{equation}
    \label{app:eq:setting:loss}
    \mathcal{L}(\theta) \coloneqq D_{\textrm{KL}}(q_\theta||p) = \mathbb{E}_{x\sim q_{\theta}}\left[\log q_{\theta}(x)\right] - \mathbb{E}_{x\sim q_{\theta}} \left[\log p(x)\right]
\end{equation}
where the target distribution $p(x)$ is given by a 2-Gaussian mixture model:
\begin{align}
    p(x) = w_{\star} \mathcal{N}(x|\mu_{\star}, I_{d}) + (1-w_{\star})\mathcal{N}(x|-\mu_{\star}, I_{d}),
\end{align}
with centroids $\mu_{\star}\in\mathbb{R}^{d}$ having norm $||\mu_{\star}||_{2}=R$. The variational family $q_{\theta}$ is given by a K-Gaussian mixture:
\begin{align}
    q_{\theta}(x) = \sum\limits_{i=1}^{K}w_{i}\mathcal{N}(x|\mu_{i},I_{d}),
\end{align}
with $K\geq 2$, with both the centroids $\mu_{i}\in\mathbb{R}^{d}$ and the weights $w_{i}\in [0,1]$ being trainable parameters. Therefore, $\theta=\{(\mu_{i},w_{i})\in\mathbb{R}^{d}\times [0,1]: i=1,\dots, K\}$. In particular, note that $w_{i}$ need to be normalised: $\sum_{i=1}^{K}w_{i}=1$.

For this setting, the loss function in \cref{app:eq:setting:loss} can be explicitly written as
\begin{equation}
    \mathcal{L}(\theta) = \sum_{i = 1}^{K} w_i \mathbb{E}_{z\sim\mathcal{N}(0,I_{d})}\left[\log\sum_{j=1}^{K} w_j \mathcal{N}(z|\mu_j - \mu_i, I_{d}) - \log\left[w_{*} \mathcal{N}(z|\mu_{\star} - \mu_i)+(1-w_{*}) \mathcal{N}(z|\mu_{\star} + \mu_i, I_{d})\right]\right]
\end{equation}
Noting that:
\begin{equation}
    \mathcal{N}(z|\mu_j - \mu_i, I_{d}) = \frac{1}{(2\pi)^{d/2}}e^{-\frac{1}{2}||z-(\mu_j - \mu_i)||^2}= \frac{1}{(2\pi)^{d/2}}e^{-\frac{1}{2}||z+\mu_i||^2 + \mu_j^{\top}(z + \mu_i) - \frac{1}{2}||\mu_j||_{2}^2}
\end{equation}
We can factor out $\exp(-||z + \mu_i||^2 / 2) / (2\pi)^{d/2}$ through logarithm to simplify the loss:
\begin{equation}
    \label{eq:app:overparam:loss}
    \mathcal{L}(\theta) = \sum_{i=1}^{K} w_i\mathbb{E}_{z\sim\mathcal{N}(0,I_{d})}\left[\log\sum_{j=1}^{K}w_j e^{\mu_j^{\top}(z + \mu_i) - \frac{1}{2}||\mu_j||_{2}^{2}} - \log\left[w_{\star} e^{\mu_{\star}^{\top}(z + \mu_{i}) - \frac{1}{2}||\mu_{\star}||_{2}^2}+(1-w_{\star}) e^{-\mu_{\star}^{\top}(z + \mu_{i}) - \frac{1}{2}||\mu_{\star}||_{2}^2}\right]\right]
\end{equation}
Note that since $z\sim\mathcal{N}(0,I_{d})$, the loss function above only depends on the vectors $\mu_{c},\mu_{\star}$ through the following real-valued summary statistics:
\begin{align}
   R^{2}=||\mu_{\star}||^{2}_{2},\qquad m_{i} = \frac{\mu_{i}^{\top}\mu_{\star}}{R^{2}}, \qquad s_{ij} = \frac{\mu_{i}^{\top}\mu_{j}}{R^{2}}, \qquad i,j=1,\dots, K
\end{align}
This observation suggests that to characterize the landscape $\mathcal{L}$, it is enough to look at its dependence on these scalar variables. This is the key idea in the analysis that will follow. To alleaviate the discussion, we will consider two simplifications in what follows:
\begin{enumerate}
    \item First, we will focus the derivation to the case $K=2$. The extension to $K>2$ is cumbersome but straightforward, and we will give the equations in \Cref{sec:app:Kg2}.
    \item Second, we fix the norm of the variational models centroids to the target norm: $||\mu_{i}||_{2}^{2} = R$. Under this assumption, the loss can be further simplified by factoring out $||\mu_j||^{2}_{2}=||\mu_{\star}||^{2}_{2}=R^{2}$:
\begin{equation}
    \label{eq:app:overparam:loss-simpl}
    \mathcal{L}(\theta) = \sum_{i=1}^{K} w_i\mathbb{E}_{z\sim\mathcal{N}(0,I_{d})}\left[\log\sum_{j=1}^{K} w_j e^{\mu_j^{\top}(z + \mu_i)} - \log\left[ w_{\star} e^{\mu_{\star}^{\top}(z + \mu_i)}+(1-w_{\star}) e^{-\mu_{\star}^{\top}(z + \mu_i)}\right]\right]
\end{equation}

    In terms of the summary statistics, this implies that $s_{ii}=1$. Although this assumption changes the optimization landscape, we empirically observe it has no important effect in the discussion of mode collapse, which is our main goal in this manuscript. 
    For the sake of completeness, we discuss the unconstrained equations in \Cref{sec:app:euclidean}.
\end{enumerate}

\subsection{Equations for the evolution of the means}
\label{appendix:stat-evolution}
We now discuss the derivation of \cref{eq:dynamics:flow} for the evolution of the summary statistics under gradient flow. By definition, spherical gradient flow is given by:
\begin{align}
\label{eq:app:sgf:means}
    \dot{\mu}_{i}=\nabla^{\mathbb{S}}_{\mu_{i}}\mathcal{L}(\mu_{i},w_{i}) \coloneqq \left(I_{d}-\frac{\mu_{i}\mu_{i}^{\top}}{R^{2}}\right)\nabla_{\mu_{i}}\mathcal{L}(\mu_{i},w_{i}), \qquad i=1,2.
\end{align}
First, let's compute the Euclidean gradient of the loss~\ref{eq:app:overparam:loss-simpl} with respect to $\mu_{1,2}$.
\begin{align}
    \nabla_{\mu_{1}}\mathcal{L} = & w_{1}\mathbb{E}_{z\sim\mathcal{N}(0,I_{d})}\left[\frac{w_{1}(z + 2\mu_{1})e^{\mu_{1}^{\top}(z + \mu_{1})} + w_{2}\mu_{2} e^{\mu_{2}^{\top}(z + \mu_{1})}}{w_{1} e^{\mu_{1}^{\top}(z + \mu_{1})} + w_{2} e^{\mu_{2}^{\top}(z + \mu_{1})}} - \frac{\mu_{\star} w_{\star} e^{\mu_{\star}^{\top}(z + \mu_{1})} - \mu_{\star} (1 - w_{\star}) e^{-\mu_{\star}^{\top}(z + \mu_{1})}}{w_{\star} e^{\mu_{\star}^{\top}(z + \mu_{1})} + (1 - w_{\star})e^{-\mu_{\star}^{\top}(z + \mu_{1})}}\right] \notag\\
    & \qquad+w_{2}\mathbb{E}_{z\sim\mathcal{N}(0,I_{d})}\left[\frac{w_{1}(\mu_{2} + z)e^{\mu_{1}^{\top}(z + \mu_{2})}}{w_{1} e^{\mu_{1}^{\top}(z + \mu_{2})} + w_{2} e^{\mu_{2}^{\top}(z + \mu_{2})}}\right]
\end{align}
Defining the sigmoid function $\sigma(t)=(1+e^{-t})^{-1}$, we can re-rewrite:
\begin{align}
    \nabla_{\mu_{1}}\mathcal{L} = & w_{1}\mathbb{E}_{z\sim\mathcal{N}(0,I_{d})}\left[(z + 2\mu_{1})\sigma\left((\mu_{1} - \mu_{2})^{\top}(z + \mu_{1}) + \log \frac{w_{1}}{w_{2}}\right) + \mu_{2} \sigma\left((\mu_{2} - \mu_{1})^{\top}(z + \mu_{1}) + \log\frac{w_{2}}{w_{1}}\right)\right. \notag\\
    &\qquad\qquad\quad\qquad+\left. \mu_{\star}\left(1 - 2\sigma\left(2\mu_{\star}^{\top}(z + \mu_{1}) + \log\frac{w_{\star}}{1 - w_{\star}}\right)\right)\right] \notag\\
    &\qquad+ w_{2}\mathbb{E}_{z\sim\mathcal{N}(0,I_{d})}\left[(z + \mu_{2})\sigma\left((\mu_{1} - \mu_{2})^{\top}(z + \mu_{2}) + \log\frac{w_{1}}{w_{2}}\right)\right]
\end{align}
It's convenient to eliminate $z g(z)$-like terms using Stein's lemma. Since $z\sim\mathcal{N}(0, I_{d})$, we have $\mathbb{E}_z[z g(z)] = \mathbb{E}_z[\nabla g(z)]$, and:
\begin{align}
    \nabla_{\mu_{1}}\mathcal{L} = & w_{1}\mathbb{E}_{z\sim\mathcal{N}(0,I_{d})}\left[\mu_{1}\sigma\left((\mu_{1} - \mu_{2})^{\top}(z + \mu_{1}) + \log\frac{w_{1}}{w_{2}}\right)\left(3 - \sigma\left((\mu_{1} - \mu_{2})^{\top}(z + \mu_{1}) + \log\frac{w_{1}}{w_{2}}\right)\right) \right. \notag\\
    &\qquad\qquad\qquad \left. +\mu_{2}\sigma\left((\mu_{2} - \mu_{1})^{\top}(z + \mu_{1}) + \log\frac{w_{2}}{w_{1}}\right)^2 + \mu_{\star} \left(1 - 2\sigma\left(2\mu_{\star}^{\top}(z + \mu_{1}) + \log\frac{w_{\star}}{1 - w_{\star}}\right)\right)\right]  \notag\\
    &+ w_{2}\mathbb{E}_{z\sim\mathcal{N}(0,I_{d})}\left[\mu_{1}\sigma\left((\mu_{1} - \mu_{2})^{\top}(z + \mu_{2}) + \log\frac{w_{1}}{w_{2}}\right)\left(1 - \sigma\left((\mu_{1} - \mu_{2})^{\top}(z + \mu_{2}) + \log\frac{w_{1}}{w_{2}}\right)\right) \right. \notag\\
    &\qquad\qquad\qquad\qquad \left. +\mu_{2}\sigma\left((\mu_{1} - \mu_{2})^{\top}(z + \mu_{2}) + \log\frac{w_{1}}{w_{2}}\right)^2\right]
\end{align}

Projecting gradient on sphere with~\ref{eq:dynamics:sphere-grad} results in
\begin{align}
    \nabla_{\mu_{1}}^{\mathbb{S}}\mathcal{L} = & (\mu_{2} - \mu_{1} s)\mathbb{E}_{z\sim\mathcal{N}(0,I_{d})}\left[w_{1}\sigma\left((\mu_{2} - \mu_{1})^{\top}(z + \mu_{1}) + \log\frac{w_{2}}{w_{1}}\right)^2 + w_{2}\sigma\left((\mu_{1} - \mu_{2})^{\top}(z + \mu_{2}) + \log\frac{w_{1}}{w_{2}}\right)^2\right]\notag\\ 
    &+w_1(\mu_{\star} - m_1\mu_{1})\mathbb{E}_{z\sim\mathcal{N}(0,I_{d}})\left[1 - 2\sigma\left(2\mu_{\star}^{\top}(z + \mu_{1}) + \log\frac{w_{\star}}{1 - w_{\star}}\right)\right]
\end{align}
Introducing the following jointly Gaussian variables: 
\begin{align}
\lambda_{\star} = \frac{\mu_{\star}}{R}z, \qquad \lambda = \frac{\mu_{1} - \mu_{2}}{\sqrt{2R^2(1 - s)}}z
\end{align}
we can rewrite:
\begin{align}
    \nabla_{\mu_{1}}^{\mathbb{S}}\mathcal{L} = & (\mu_{2} - \mu_{1} s)\mathbb{E}_{\lambda} \left[w_{1}\sigma\left(R^2(s - 1) + \lambda R\sqrt{2(1 - s)} + \log\frac{w_{2}}{w_{1}}\right)^2 + w_{2}\sigma\left(R^2(s - 1) + \lambda R\sqrt{2(1 - s)} + \log\frac{w_{1}}{w_{2}}\right)^2\right] \notag\\
    &+ w_{1}(\mu_{\star} - m_1\mu_{1})\mathbb{E}_{\lambda_{\star}}\left[1 - 2\sigma\left(2R^2m_1 + 2R\lambda_{\star} + \log\frac{w_{\star}}{1 - w_{\star}}\right)\right]
\end{align}
Introducing the following functions:
\begin{align}
    f(s) &=  \mathbb{E}_\lambda \left[w_{1}\sigma\left(R^2(s - 1) + \lambda R\sqrt{2(1 - s)} + \log\frac{w_{2}}{w_{1}}\right)^2 +  w_{2}\sigma\left(R^2(s - 1) + \lambda R\sqrt{2(1 - s)} + \log\frac{w_{1}}{w_{2}}\right)^2\right] \\
    g(m) &= \mathbb{E}_{\lambda_{\star}} \left[1 - 2\sigma\left(2R^2m_1 + 2R\lambda_{\star} + \log\frac{w_{\star}}{1 - w_{\star}}\right)\right] 
\end{align}
We can conveniently write the spherical gradient as:
\begin{align}
    \nabla_{\mu_{1}}^\mathbb{S}\mathcal{L} &=  (\mu_{2} - \mu_{1} s) f(s) + w_{1}(\mu_{\star} - m_1\mu_{1})g(m_1)
\end{align}
Using symmetry of loss $\mu_{1}\to\mu_{2}$, $\mu_{\star}\to -\mu_{\star}$, $w_{1}\to w_{2}$, $w_{\star}\to 1 - w_{\star}$ we obtain gradient with respect to $\mu_{2}$
\begin{align}
    \nabla_{\mu_{2}}^{\mathbb{S}}\mathcal{L} = & (\mu_{1} - \mu_{2} s)f(s) + w_{2}(\mu_{\star} - m_2\mu_{2})g(m_2)
\end{align}
Finally, we compute the evolution of sufficient statistics by writing:
\begin{align}
    \dot{m}_{1,2} &=  \frac{1}{R^2}\mu_\star^{\top}\dot{\mu}_{1,2} = -\frac{1}{R^2}\mu_\star^{\top}\nabla_{\mu_{1,2}}^{\mathbb{S}}\mathcal{L} \notag\\
    &= -\left[(m_{2,1} - m_{1,2} s)f(s) + w_{1,2}(1 - m_{1,2}^2)g(m_{1,2})\right] \\
    \dot{s} &= \frac{1}{R^2}\dot{\mu}_1^{\top}\mu_{2} + \mu_{1}^{\top}\dot{m}_2 = -\frac{1}{R^2}\left[\mu_{2}^{\top}\nabla_{\mu_{1}}^{\mathbb{S}}\mathcal{L} + \mu_{1}^{\top}\nabla_{\mu_{2}}^{\mathbb{S}}\mathcal{L}\right] \notag\\
    &=-\left[2(1 - s^2)f(s) + w_{1}(m_2 - m_1s)g(m_1) + w_{2}(m_1 - m_2s)g(m_2)\right]
\end{align}
which are precisely the \cref{eq:dynamics:flow} writen in the main.

\subsection{Fixed points of the mean evolution}
\label{appendix:fixpt}
By definition, the fixed points of the flow \cref{eq:dynamics:flow} are the zero gradient points. That is, we look for $(m_1, m_2, s)\in[-1,1]^{3}$ such that:
\begin{align}
    \label{eq:appendix:flow1}
    & (m_2 - m_1 s)f(s) + w_{1}(1 - m_1^2)g(m_1) = 0\\
    \label{eq:appendix:flow2}
    & (m_1 - m_2 s)f(s) + w_{2}(1 - m_2^2)g(m_2) = 0\\
    \label{eq:appendix:flow3}
    & 2(1 - s^2)f(s) + w_{1}(m_2 - m_1 s)g(m_1) + w_{2}(m_1 - m_2 s)g(m_2) = 0
\end{align}
First, suppose that $1 - m_1^2 = 0$. Then since $f(s) > 0$ for any $s$, from \cref{eq:appendix:flow1} we get $m_2 - m_1 s = 0$. \Cref{eq:appendix:flow2} simplifies to 
\begin{equation}
    m_1(1 - s^2)f(s) + w_{2}(1 - s^2)g(m_2) = 0
\end{equation}
Substituting into \cref{eq:appendix:flow3} gives 
\begin{equation}
    2(1 - s^2)f(s) + w_{2}m_1(1 - s^2)g(m_2) = 2(1 - s^2)f(s) - m_1^2(1-s^2)f(s) = (1 - s^2) f(s) = 0
\end{equation}
Again by positivity of $f(s)$, $s = \pm 1$. By symmetry, if either $m_1$ or $m_2$ is $\pm 1$, then $s = \pm 1$ and we have global minimum/flipped fixed point at $s=-1$ or full alignment at $s=1$.

Second, suppose that both $m_{1,2}\in(-1, 1)$, but $g(m_1)=0$. Then \cref{eq:appendix:flow1} implies $m_2 - m_1 s = 0$ and \cref{eq:appendix:flow3} simplifies to
\begin{equation}
    2(1 - s^2)f(s)^2 = w_{2}^2(1 - m_1^2 s^2)m_1^2(1 - s^2)^2\frac{f(s)^2}{w_{2}^2(1 - m_1^2 s^2)^2}\Rightarrow 2 = \frac{m_1^2(1 - s^2)}{1 - m_1^2s^2} \lor 1 - s^2 = 0
\end{equation}
The first equation results in the extremal solution $m_1^2 = 1$,  which we have already considered. The second one gives $s = \pm 1$. If $s = 1$, then $m_2 = m_1 = m$, where $g(m) = 0$, which we check to be a full alignment fixed point. We would reach the same conclusion if we started from $g(m_2) = 0$.

Third, consider the general case when $m_{1,2}\in(-1,1)$ and $g(m) \neq 0$. Unfortunately, an explicit solution for the general fixed points is not available. Instead, we derive an unexpected geometrical connection between the three summary statistics. 

Consider the $3\times 3$ matrix $P$ formed by scalar products of means
\begin{equation}
    P = \frac{1}{R^2}\begin{bmatrix}
        \mu_{\star}^{\top}\mu_{\star} & \mu_{1}^{\top}\mu_{\star} & \mu_{2}^{\top}\mu_{\star} \\
        \mu_{\star}^{\top}\mu_{1} & \mu_{1}^{\top}\mu_{1} & \mu_{2}^{\top}\mu_{1} \\
        \mu_{\star}^{\top}\mu_{2} & \mu_{1}^{\top}\mu_{2} & \mu_{2}^{\top}\mu_{2}
    \end{bmatrix} = \begin{bmatrix}
        1 & m_1 & m_2 \\
        m_1 & 1 & s \\
        m_2 & s & 1
    \end{bmatrix}
\end{equation}
It is semidefinite positive because for any vector $v\in \mathbb{R}^3$, 
\begin{equation}
    v^{\top}Pv = \sum_{ij} v_i\mu_i^{\top}\mu_j v_j = \left(\sum_i v_i\mu_i\right)^{\top}\left(\sum_j v_j\mu_j\right) \geq 0
\end{equation} 
Any semidefinite matrix has a non-negative determinant. Let us study how this condition evolves throughout the flow \cref{eq:dynamics:flow} by writing:
\begin{align}
    \frac{\rm d}{\rm{d}t}\det P &= \frac{\rm d}{\rm{d}t}(1 + 2m_1m_2s - m_1^2 - m_2^2 - s^2) \notag\\
    &= 2\left[\dot{m}_1(m_2s - m_1) + \dot{m}_2(m_1s - m_2) + \dot{s}(m_1m_2 - s)\right] \notag\\
    &= 2 (-1 + m_1^2 + m_2^2 - 2 m_1 m_2 s + s^2)\left[2 s f(s) + m_1 w_{1} g(m_1) + m_2 w_{2} g(m_2))\right]\notag\\
    &=-2 \left[2 s f(s) + m_1 w_{1} g(m_1) + m_2 w_{2} g(m_2))\right]\det P
\end{align}
At the fixed point, the time derivative of $\det P$ is zero, meaning that either $\det P = 0$ or the expression in the brackets vanishes. Let us start from the second case. Multiplying it by $s$ and summing with \cref{eq:appendix:flow3} we get:
\begin{equation}
    2f(s) + w_{1}m_2g(m_1) + w_{2}m_1g(m_2) = 0
\end{equation}
We can substitute $f(s)$ into \cref{eq:appendix:flow1} and \cref{eq:appendix:flow2} and rearrange terms to get:
\begin{align}
    & w_{1} g(m_1)(2 - 2m_1^2 - m_2^2 + m_1m_2s) = w_{2} m_1 g(m_2)(m_2 - m_1 s) \\
    & w_{2} g(m_2)(2 - 2m_2^2 - m_1^2 + m_1m_2s) = w_{1} m_2 g(m_1)(m_1 - m_2 s)
\end{align}
Since we have assumed earlier that $g(m_1) \neq 0$ and $g(m_2)\neq 0$ we can combine these expressions into:
\begin{equation}
    m_1(m_2 - m_1 s)(2 - 2m_2^2 - m_1^2 + m_1m_2s) = m_2(m_1 - m_2 s)(2 - 2m_1^2 - m_2^2 + m_1m_2s)
\end{equation}
Since again $g(m_1)\neq 0$, $1 - m_1^2\neq 0$, $f(s) > 0$, $m_2 - m_1 s \neq 0$ and by analogy $m_1 - m_2 s \neq 0$. Since the weights are unbalanced we can't have $m_{1,2} = 0$ either, so the only possibility is that $2 - 2m_2^2 - m_1^2 + m_1m_2s = 2 - 2m_1^2 - m_2^2 + m_1m_2s = 0$, implying that $\det P$ equal to half sum of these terms is also zero. Therefore, in both the first and second cases:
\begin{equation}
    \det P = 1 + 2m_1m_2s - m_1^2 - m_2^2 - s^2 =  0\Rightarrow s = m_1m_2 \pm \sqrt{(1 - m_1^2)(1 - m_2^2)}
\end{equation}
This interesting geometrical constraint can be used to restrict the space of possible fixed points.

\paragraph{Hessian calculation}
To investigate the stability of the fixed points we need to linearize the flow equations \cref{eq:dynamics:flow} around them. The linear system is written as
\begin{equation}
    \frac{d}{dt}\begin{bmatrix}
        \delta m_1\\
        \delta m_2\\
        \delta s
    \end{bmatrix} = -H\begin{bmatrix}
        \delta m_1\\
        \delta m_2\\
        \delta s
    \end{bmatrix}
\end{equation}
where we call $H$ as Hessian at the fixed point.

First, when $m_1=\pm 1$ and $s=\pm 1$, $m_2 - m_1 s$ and $1 - m_1^2$ terms vanish and we can discard terms containing derivatives of $f$ and $g$ arising during linearization
\begin{align}
    & \delta\dot{m}_1 = -\left[(\delta m_2 - s\delta m_1 - m_1\delta s)f(s) - \frac{2\gamma m_1}{\gamma + 1}g(m_1)\delta m_1\right] \\
    & \delta\dot{m}_2 = -\left[(\delta m_1 - s\delta m_2 - m_2\delta s)f(s) - \frac{2m_2}{\gamma + 1}g(m_2)\delta m_2\right] \\
    & \delta\dot{s} = -\left[-4sf(s)\delta s + \frac{\gamma}{\gamma + 1}g(m_1)(\delta m_2 - s\delta m_1 - m_1\delta s) + \frac{1}{\gamma + 1}g(m_2)(\delta m_1 - s\delta m_2 - m_2\delta s)\right]
\end{align}

Eigenvalues of $H$ at global minimum $(m^*_1, m^*_2, s^*) = (1, -1, -1)$ are
\begin{align}
    & \lambda = 2 f(-1) - w_1 g(1) + w_2 g(-1) \\
    & \lambda_\pm = 2 f(-1) - w_1 g(1) + w_2 g(-1)\pm\sqrt{4f(-1)^2 + (w_1 g(1) + w_2 g(-1))^2}
\end{align}

At flipped fixed point $(m^*_1, m^*_2, s^*) = (-1, 1, -1)$ they are
\begin{align}
    & \lambda = 2 f(-1) - w_2 g(1) + w_1 g(-1) \\
    & \lambda_\pm = 2 f(-1)  - w_2 g(1) + w_1 g(-1)\pm \sqrt{4f(-1)^2 + (w_2 g(1) + w_1 g(-1))^2}
\end{align}

At perfect alignment mode collapse fixed point $(m^*_1, m^*_2, s^*) = (1, 1, 1)$ they are
\begin{align}
    & \lambda = -2f(1) - g(1) \\
    & \lambda_\pm = -2f(1) - g(1)\pm \sqrt{4f(1)^2 + \frac{(1 - w_2)^2}{(1 + w_2)^2} g(1)^2}
\end{align}

At opposite fixed point $(m^*_1, m^*_2, s^*) = (-1, -1, 1)$ they are
\begin{align}
    & \lambda = -2f(1) + g(-1) \\
    & \lambda_\pm = -2f(1) + g(-1)\pm \sqrt{4f(1)^2 + \frac{(1 - w_2)^2}{(1 + w_2)^2} g(-1)^2}
\end{align}

For the last perfect alignment fixed point, $m_2 = m_1 = m$, where $g(m) = 0$ and $s = 1$. Then $m_2 - m_1 s$ vanishes and linearized equations are
\begin{align}
    & \delta\dot{m}_1 = -\left[(\delta m_2 - s\delta m_1 - m_1\delta s)f(s) + \frac{\gamma}{\gamma + 1}(1 - m_1^2)g'(m_1)\delta m_1\right]\\
    & \delta\dot{m}_2 = -\left[(\delta m_1 - s\delta m_2 - m_2\delta s)f(s) + \frac{1}{\gamma + 1}(1 - m_2^2)g'(m_2)\delta m_2\right] \\
    & \delta\dot{s} = 4sf(s)\delta s
\end{align}

Eigenvalues of $H$ are
\begin{align}
    & \lambda = -4f(1) \\
    & \lambda_\pm  = w_1(1 - m^2)g'(m) - 1\pm \sqrt{f(1)^2 + \frac{1}{4}w_1^2(1 - m^2)^2g'(m)^2}
\end{align}
For any $w_{1,2}$ and $R$, the point is unstable due to $\lambda < 0$, which is expected. For other points above we numerically confirm that perfect alignment fixed points are unstable, global minimum is stable, and flipped fixed point is stable for large enough $R$ and is unstable otherwise.

\subsection{Weights evolution}
\label{appendix:weight-evolution}
We now turn our attention to the evolution of the weights $w_{1},w_{2}\in[0,1]$. Note that the naive flow $\dot{w}_{i} = \nabla_{w_{i}}\mathcal{L}(\theta)$ does not respect the normalization constraint $w_{1}+w_{2}= 1$, and therefore take us out of the space of probability density functions. Different solutions for this can be considered, each leading to a different VI algorithm with different optimization properties. Here, we consider two particular solutions: a \emph{reparametrized flow} and a \emph{projected flow}, which we discuss below. In both cases, the key idea is to see the flow as the vanishing step-size limit of a gradient descent algorithm that preserves the weights.

\subsubsection{Reparametrized flow}
\label{sec:app:rep}
The first idea consists of enforcing the normalisation by considering the following reparametrization of the weights:
\begin{equation}
    \label{eq:weight:param}
    w_{1} = \frac{v_1}{v_1 + v_2}, \quad w_{2} = \frac{v_2}{v_1 + v_2};
\end{equation}
where $v_{1},v_{2}\in\mathbb{R}_{+}$. This can be implemented via the following gradient descent scheme:
\begin{enumerate}
    \item Initialize differentiable parameters $v_1 = w_1, v_2 = w_2$
    \item Parameterize weights as
    \begin{equation}
        \label{eq:weight:param}
        w_1 = \frac{v_1}{v_1 + v_2};\quad w_2 = \frac{v_2}{v_1 + v_2};
    \end{equation}
    \item Update parameters with gradient descent
    \begin{equation}
        v_1' = v_1 - \eta\frac{\partial\mathcal{L}(w_1, w_2)}{\partial v_1};\quad v_2' = v_2 - \eta\frac{\partial\mathcal{L}(w_1, w_2)}{\partial v_2};
    \end{equation}
    Here $w_{1,2}$ are functions of $v_{1,2}$ according to~\ref{eq:weight:param}.
    \item Obtain new weights with parameterization~\cref{eq:weight:param}.
    \begin{equation}
        \label{eq:weight:param-proj}
        w_1' = \frac{v_1'}{v_1' + v_2'};\quad w_2' = \frac{v_2'}{v_1' + v_2'};
    \end{equation}
    Thus, we have explicitly preserved normalization constraint at the end of each iteration.
\end{enumerate}
It is easy to see that this gradient descent scheme preserve the normalization $w_{1}+w_{2}=1$. To implement the above, we need to compute the gradients in $v_{1},v_{2}$. First, consider the gradient of the loss with respect to $v_{1}$, it is given by:
\begin{align}
    \frac{\partial\mathcal{L}}{\partial v_1} &= \frac{\partial\mathcal{L}}{\partial w_1}\frac{\partial w_1}{\partial v_1} + \frac{\partial\mathcal{L}}{\partial w_2}\frac{\partial w_2}{\partial v_1} = \frac{v_2}{(v_1 + v_2)^2}\frac{\partial\mathcal{L}}{\partial w_1} -  \frac{v_2}{(v_1 + v_2)^2}\frac{\partial\mathcal{L}}{\partial w_2} \notag\\
    &= w_2\left(\frac{\partial\mathcal{L}}{\partial w_1} - \frac{\partial\mathcal{L}}{\partial w_2}\right)
\end{align}
Similarly, by symmetry the gradient of the loss with respect to $v_{2}$ is given by: 
\begin{equation}
    \frac{\partial\mathcal{L}}{\partial v_2} = w_1\left(\frac{\partial\mathcal{L}}{\partial w_2} - \frac{\partial\mathcal{L}}{\partial w_1}\right)
\end{equation}
Therefore, gradient descent update (step 4, \cref{eq:weight:param-proj}) is given by:
\begin{align}
    w_1' &= w_1 - \eta\left[v_2\frac{\partial\mathcal{L}}{\partial v_1} - v_1\frac{\partial\mathcal{L}}{\partial v_2}\right] + O(\eta^2) \\
    &= w_1 - \eta(w_1^2 + w_2^2)\left(\frac{\partial\mathcal{L}}{\partial w_1} - \frac{\partial\mathcal{L}}{\partial w_2}\right) + O(\eta^2)
\end{align}
And hence the limiting $\eta\to 0^{+}$ flow is:
\begin{align}
\label{eq:app:flow:rep}
    & \dot{w}_1 = -(w_1^2 + w_2^2)\left(\frac{\partial\mathcal{L}}{\partial w_1} - \frac{\partial\mathcal{L}}{\partial w_2}\right) \\
    & \dot{w_2} = -\dot{w}_1
\end{align}
Finally, it remains to show that the equations above can be written in terms of the summary statistics $(m_{1},m_{2},s)$. For that, we need to compute the gradients of the loss with respect to the weights $(w_{1},w_{2})$. Starting from $w_{1}$:
\begin{align}
    \frac{\partial \mathcal{L}}{\partial w_{1}} &= \mathbb{E}_{z\sim\mathcal{N}(0,I_{d})}\left[\frac{w_{1} e^{\mu_{1}^{\top}(z + \mu_{1})}}{w_{1} e^{\mu_{1}^{\top}(z + \mu_{1})} + w_{2} e^{\mu_{2}^{\top}(z + \mu_{1})}} + \frac{w_{2} e^{\mu_{1}^{\top}(z + \mu_{2})}}{w_{1} e^{\mu_{1}^{\top} (z + \mu_{2})} + w_{2} e^{\mu_{2}^{\top} (z + \mu_{2})}}\right] \notag\\
    &\quad+\mathbb{E}_{z\sim\mathcal{N}(0,I_{d})}\left[\log\left(w_{1} e^{\mu_{1}^{\top}(z + \mu_{1})} + w_{2} e^{\mu_{2}^{\top}(z + \mu_{1})}\right) - \log\left(w_{\star} e^{\mu_{\star}^{\top}(z + \mu_{1})} + (1 - w_{\star})e^{-\mu_{\star}^{\top}(z + \mu_{1})}\right)\right] \\
    &= \mathbb{E}_{z\sim\mathcal{N}(0,I_{d})}\left[\sigma\left(\mu_{1}^{\top}(\mu_{1} - \mu_{2}) + (\mu_{1} - \mu_{2})^{\top}z + \log\frac{w_{1}}{w_{2}}\right) + \frac{w_{2}}{w_{1}}\sigma\left(\mu_{2}^{\top}(\mu_{1} - \mu_{2}) + (\mu_{1} - \mu_{2})^{\top}z + \log\frac{w_{1}}{w_{2}}\right)\right] \notag\\
    &\quad+\mathbb{E}_{z\sim\mathcal{N}(0,I_{d})}\left[\log\frac{w_{1}}{w_{\star}} + \mu_{1}^{\top}(z + \mu_{1}) - \mu_{\star}^{\top}(z + \mu_{1}) - \log\sigma\left((\mu_{1} - \mu_{2})^{\top}(z + \mu_{1}) + \log\frac{w_{1}}{w_{2}}\right)  \right]\notag\\  
    &\quad+\mathbb{E}_{z\sim\mathcal{N}(0,I_{d})}\left[\log\sigma\left(2\mu_{\star}^{\top}(z + \mu_{1}) + \log\frac{w_{\star}}{1 - w_{\star}}\right)\right]
\end{align}
As in \Cref{appendix:stat-evolution}, we now introduce the jointly Gaussian random variables: 
\begin{align}
\lambda = \frac{\mu_{1} - \mu_{2}}{\sqrt{2R^2(1-s)}}z, \qquad\lambda_{\star} = \frac{\mu_{\star}}{R}z.    
\end{align}
Introduce sufficient statistics and rewrite:
\begin{align}
    \label{eq:appendix:dldw1}
    \frac{\partial \mathcal{L}}{\partial w_{1}} = &\mathbb{E}_{\lambda}\left[\sigma\left(R^2(1 - s) + \lambda R\sqrt{2(1 - s)} + \log\frac{w_{1}}{w_{2}}\right) + \frac{w_{2}}{w_{1}}\sigma\left(-R^2(1 - s) + \lambda R\sqrt{2(1 - s)} + \log\frac{w_{1}}{w_{2}}\right)\right]\notag\\ 
    &-\mathbb{E}_{\lambda,\lambda_{\star}}\left[\log\sigma\left(R^2(1 - s) + \lambda R\sqrt{2(1 - s)} + \log\frac{w_{1}}{w_{2}}\right) + \log\sigma\left(2R^2m_1 + 2R\lambda_{\star} + \log\frac{w_{\star}}{1 - w_{\star}}\right)\right]\notag\\
    &+ \log \frac{w_{1}}{w_{\star}} + R^2(1 - m_1) 
\end{align}

We can obtain derivative $\partial \mathcal{L}/\partial w_{2}$ in analogy by exploiting symmetry of loss under $\mu_{1}\to\mu_{2}$, $\mu_{\star}\to-\mu_{\star}$, $w_{1}\to 1 - w_{1} = w_{2}$, $w_{\star}\to 1 - w_{\star}$, which acts on statistics as $m_1\to -m_2$, $s\to s$, $\lambda\to-\lambda$ and $\lambda_{\star}\to-\lambda_{\star}$. We obtain
\begin{align}
    \label{eq:appendix:dldw2}
    \frac{\partial\mathcal{L}}{\partial w_{2}} = &\mathbb{E}_{\lambda}\left[\sigma\left(R^2(1 - s) - \lambda R\sqrt{2(1-s)} - \log\frac{w_{1}}{w_{2}}\right)+\frac{w_{1}}{w_{2}}\sigma\left(-R^2(1 - s) - \lambda R\sqrt{2(1 - s)} - \log\frac{w_{1}}{w_{2}}\right)\right]  \notag\\
    &-\mathbb{E}_{\lambda,\lambda_{\star}}\left[\log\sigma\left(R^2(1 - s) - \lambda R\sqrt{2(1 - s)} - \log\frac{w_{1}}{w_{2}}\right) + \log\sigma\left(-2R^2m_2 - 2R\lambda_{\star} - \log\frac{w_{\star}}{1 - w_{\star}}\right)\right] \notag\\
    &+ \log \frac{w_{2}}{1 - w_{\star}} + R^2(1 + m_2)
\end{align}
Together with \cref{eq:app:flow:rep}, this shows that the flow of the weights can be written in terms of $(w_{1},w_{2},m_{1},m_{2},s)$.

\subsubsection{Projected gradient}
\label{appendix:proj-grad}
An alternative approach to reparametrizing the variables consist of projecting the gradient on the weights back in the constraint set, similar to the spherical flow for the means. This can be implemented as follows:
\begin{enumerate}
    \item At every step, we take a standard gradient descent on the weights: 
    \begin{equation}
        w_{1}' = w_{1} - \eta\frac{\partial\mathcal{L}}{\partial w_{1}};\quad w_{2}' = w_{2} - \eta\frac{\partial\mathcal{L}}{\partial w_{2}};
    \end{equation}
    \item Next, we normalize the weights such that $w_{1}+w_{2}=1$:
    \begin{equation}
        \label{eq:weight:grad-proj}
        w_{1}'' = \frac{w_{1}'}{w_{1}' + w_{2}'};\quad w_{2}'' = \frac{w_{2}'}{w_{1}' + w_{2}'};
    \end{equation}
\end{enumerate}
To evaluate the update, we need to compute:
\begin{align}
    w_{1}'' &= \frac{w_{1} - \eta \partial \mathcal{L} / \partial w_{1}}{w_{1} - \eta \partial \mathcal{L} / \partial w_{1} + w_{2} - \eta \partial \mathcal{L} / \partial w_{2}} \notag\\
    &= \left(w_{1} - \eta\frac{\partial \mathcal{L}}{\partial w_{1}}\right)\left(1 + \eta\frac{\partial \mathcal{L}}{\partial w_{1}} + \eta\frac{\partial \mathcal{L}}{\partial w_{2}}\right)+O(\eta^{2}) \notag\\
    &=w_{1} - \eta\left[(1 - w_{1})\frac{\partial \mathcal{L}}{\partial w_{1}} - w_{1}\frac{\partial \mathcal{L}}{\partial w_{2}}\right]+O(\eta^{2})\\ 
    &= w_{1} - \eta \left[w_{2}\frac{\partial \mathcal{L}}{\partial w_{1}} - w_{1}\frac{\partial \mathcal{L}}{\partial w_{2}}\right]+O(\eta^{2})
\end{align}
where in the second equality we expanded in small $\eta\ll 1$, keeping only the first order terms. Therefore, if we want a gradient flow to mimic the projected gradient, we have to compute $w_{2}\partial \mathcal{L}/\partial w_{1} - w_{1}\partial \mathcal{L}/\partial w_{2}$ instead of $\partial \mathcal{L}/\partial w_{1}$. Together with the symmetric expression for $w_{2}''$ and the gradient of the weights previously derived in \cref{eq:appendix:dldw1} and \cref{eq:appendix:dldw2}, this allows to write the projected flow in terms of the summary statistics.

However, we find that in comparison with the reparametrized flow from \cref{sec:app:rep}, this projected flow is more prompt to mode collapse with one of the weights going quickly to zero, in particular at small $R$. Let us consider limit $R\to 0^{+}$ and compute the gradients in \cref{eq:appendix:dldw1} and \cref{eq:appendix:dldw2} up to second order of $R$:
\begin{align}
    \frac{\partial\mathcal{L}}{\partial w_{1}} = &\mathbb{E}_{\lambda} \left[w_{1} + w_{1}w_{2}\lambda R\sqrt{2(1 - s)} + (1 - s)w_{1}w_{2}(w_{1}(1 - \lambda^2) + w_{2}(1 + \lambda^2))R^2\right] \notag\\
    &+\mathbb{E}_{\lambda}\left[w_{2} + w_{2}^2\lambda R\sqrt{2(1 - s)} - (1 - s)w_{2}^2(w_{1}(1 + \lambda^2) + w_{2}(1 - \lambda^2))R^2\right] \notag\\
    &-\mathbb{E}_{\lambda,\lambda_{\star}}\left[\log w_{1} + w_{2}\lambda R\sqrt{2(1 - s)} + (1 - s)w_{2}(w_{2} + w_{1}(1 - \lambda^2))R^2\right]  \notag\\
    &+\mathbb{E}_{\lambda,\lambda_{\star}}\left[\log w_{\star} + 2\lambda R(1 - w_{\star}) - 2(1 - w_{\star})(w_{\star}\lambda^2 - m_1)R^2\right] \notag\\
    & +\log \frac{w_{1}}{w_{\star}} + R^2(1 - m_1^2) + o(R^2)
\end{align}
After averaging out and using symmetry we get $\partial \mathcal{L}/\partial w_{1,2}$
\begin{align}
    & \frac{\partial \mathcal{L}}{\partial w_{1}} = 1 + R^2((1 - m_1) - (1 - s)w_{2}^2 + 2(1 - w_{\star})(m_1 - w_{\star}))\\
    & \frac{\partial \mathcal{L}}{\partial w_{2}} = 1 + R^2((1 + m_2) - (1 - s)w_{1}^2 + 2w_{\star}(-m_2 - 1 + w_{\star}))
\end{align}
At the gradient flow level, at $R\to 0^{+}$, $w_{1}$ evolves as:
\begin{equation}
    \dot{w_{1}} = -\left(w_{2}\frac{\partial l}{\partial w_{1}} - w_{1}\frac{\partial l}{\partial w_{2}}\right) = -(1 - w_{1} - w_{1}) = 2(w_{1} - 1 / 2)
\end{equation}
The flow is linear and has an unstable expansion factor $2$. A non-zero radius gives a $O(R^2)$ correction which is insufficient to make the flow stable up until a certain radius threshold. Thus, $w_{1}$ is just shot into $0$ or $1$, depending on initialization.

\subsection{Generalizations}
In this section, we briefly discuss the generalization of the flow equations derived in \Cref{appendix:stat-evolution} to more general cases.

\subsubsection{General K-Gaussian mixture}
\label{sec:app:Kg2}

Let us start with loss~\ref{eq:app:overparam:loss}, use that $\|\mu_\star\|^2 = R^2$ to simplify exponents and rewrite summation index $i\to k$ to obtain
\begin{equation}
    \label{eq:app:overparam:loss-rew}
    \mathcal{L}(\theta) = \sum_{k=1}^{K} w_k\mathbb{E}_{z\sim\mathcal{N}(0,I_{d})}\left[\log\sum_{j=1}^{K}w_j e^{\mu_j^{\top}(z + \mu_k) + \frac{1}{2}(R^2 - ||\mu_j||_{2}^{2})} - \log\left[w_{\star} e^{\mu_{\star}^{\top}(z + \mu_{k})}+(1-w_{\star}) e^{-\mu_{\star}^{\top}(z + \mu_{k})}\right]\right]
\end{equation}
The gradient with respect to mean $\mu_i$ is then
\begin{align}
    \nabla_{\mu_i}\mathcal{L} = 
    & \sum_{k=1}^K w_k\mathbb{E}_z \frac{\sum_{j=1}^K w_j \left[(z + \mu_k - \mu_j)\delta_{ij} + \mu_j \delta_{ik}\right]e^{\mu_j^{\top}(z + \mu_k) + \frac{1}{2}(R^2 - ||\mu_j||_{2}^{2})}}{\sum_{j=1}^{K}w_j e^{\mu_j^{\top}(z + \mu_k) + \frac{1}{2}(R^2 - ||\mu_j||_{2}^{2})}} + \notag\\
    & w_i\mathbb{E}_z\mu_\star\left(1 - 2\sigma\left(2\mu_\star^{\top}(z + \mu_i) \log\frac{w_\star}{1 - w_\star}\right)\right) = \notag\\
    & \sum_{k=1}^K w_k \mathbb{E}_z\frac{w_i(z + \mu_k - \mu_i)e^{\mu_i^{\top}(z + \mu_k) + \frac{1}{2}(R^2 - ||\mu_i||_{2}^{2})}}{\sum_{j=1}^K w_j e^{\mu_j^{\top}(z + \mu_k) + \frac{1}{2}(R^2 - ||\mu_j||_{2}^{2})}} + w_i\mathbb{E}_z\frac{\sum_{j=1}^K w_j\mu_j e^{\mu_j^{\top}(z + \mu_i) + \frac{1}{2}(R^2 - \|\mu_j\|^2_2)}}{\sum_{j=1}^K w_j e^{\mu_j^{\top}(z + \mu_i) + \frac{1}{2}(R^2 - \|\mu_j\|^2_2)}} + \notag\\
    & w_i\mathbb{E}_z\mu_\star\left(1 - 2\sigma\left(2\mu_\star^{\top}(z + \mu_i) + \log\frac{w_\star}{1 - w_\star}\right)\right)
\end{align}
Eliminate $zg(z)$-like terms with Stein's lemma
\begin{align}
    \nabla_{\mu_i}\mathcal{L} = & \sum_{k=1}^K w_k \mathbb{E}_z\frac{w_i(\mu_k - \mu_i)e^{\mu_i^{\top}(z + \mu_k) + \frac{1}{2}(R^2 - ||\mu_i||_{2}^{2})}}{\sum_{j=1}^K w_j e^{\mu_j^{\top}(z + \mu_k) + \frac{1}{2}(R^2 - ||\mu_j||_{2}^{2})}} + w_i\mu_i\sum_{k=1}^K w_k\mathbb{E}_z\frac{ e^{\mu_i^{\top}(z + \mu_k) + \frac{1}{2}(R^2 - ||\mu_i||_{2}^{2})}}{\sum_{j=1}^K w_j e^{\mu_j^{\top}(z + \mu_k) + \frac{1}{2}(R^2 - ||\mu_j||_{2}^{2})}} \notag\\
    - & w_i\sum_{k=1}^K w_k\mathbb{E}_z\frac{e^{\mu_i^{\top}(z + \mu_k) + \frac{1}{2}(R^2 - ||\mu_i||_{2}^{2})}\sum_{j=1}^K w_j\mu_je^{\mu_j^{\top}(z + \mu_k) + \frac{1}{2}(R^2 - ||\mu_j||_{2}^{2})}}{\left(\sum_{j=1}^K w_j e^{\mu_j^{\top}(z + \mu_k) + \frac{1}{2}(R^2 - ||\mu_j||_{2}^{2})}\right)^2} \notag\\
    + & w_i\mathbb{E}_z\frac{\sum_{j=1}^K w_j\mu_j e^{\mu_j^{\top}(z + \mu_i) + \frac{1}{2}(R^2 - \|\mu_j\|^2_2)}}{\sum_{j=1}^K w_j e^{\mu_j^{\top}(z + \mu_i) + \frac{1}{2}(R^2 - \|\mu_j\|^2_2)}} + w_i\mathbb{E}_z\mu_\star\left(1 - 2\sigma\left(2\mu_\star^{\top}(z + \mu_i) + \log\frac{w_\star}{1 - w_\star}\right)\right)
\end{align}

\subsubsection{Euclidean flow}
\label{sec:app:euclidean}

Let's compute the Euclidean gradient of the loss~\cref{eq:app:overparam:loss-rew} when $K=2$
\begin{align}
    \nabla_{\mu_1}\mathcal{L} = & w_1\mathbb{E}_z\frac{w_1(z + \mu_1)e^{\mu_1^{\top}(z + \mu_1) + \frac{1}{2}(R^2 - \|\mu_1\|^2_2)} + w_2 \mu_2 e^{\mu_2^{\top}(z+\mu_1) + \frac{1}{2}(R^2 - \|\mu_2\|^2_2)}}{w_1 e^{\mu_1^{\top}(z + \mu_1) + \frac{1}{2}(R^2 - \|\mu_1\|^2_2)} + w_2 e^{\mu_2^{\top}(z+\mu_1) + \frac{1}{2}(R^2 - \|\mu_2\|^2_2)}} - \notag\\
    & w_1\mathbb{E}_z\frac{\mu_\star w_\star e^{\mu_\star^{\top}(z + \mu_1)} - \mu_\star(1 - w_\star)e^{-\mu_\star^{\top}(z + \mu_1)}}{w_\star e^{\mu_\star^{\top}(z + \mu_1)} + (1 - w_\star)e^{-\mu_\star^{\top}(z + \mu_1)}} + \notag\\
    & w_2\mathbb{E}_z \left[\frac{w_1(z + \mu_2 - \mu_1)e^{\mu_1^{\top}(z + \mu_2) + \frac{1}{2}(R^2 - \|\mu_1\|^2)_2}}{w_1e^{\mu_1^{\top}(z + \mu_2) + \frac{1}{2}(R^2 - \|\mu_1\|^2_2)} + w_2e^{\mu_2^{\top}(z + \mu_2) + \frac{1}{2}(R^2 - \|\mu_2\|^2_2)}}\right]
\end{align}

Rewrite with sigmoid function as
\begin{align}
    \nabla_{\mu_1}\mathcal{L} = & w_1 \mathbb{E}_z\left[(z + \mu_1)\sigma\left((\mu_1-\mu_2)^{\top}(z + \mu_1) + \frac{1}{2}(\|\mu_2\|^2_2 - \|\mu_1\|^2_2) + \log\frac{w_1}{w_2}\right)\right.\notag \\
    & + \left. \mu_2\sigma\left((\mu_2 - \mu_1)^{\top}(z + \mu_1) + \frac{1}{2}(\|\mu_1\|^2_2 - \|\mu_2\|^2_2) + \log\frac{w_2}{w_1}\right)\right.\notag\\
    & + \left. \mu_\star\left(1 - 2\sigma\left(2\mu_\star^{\top}(z + \mu_1) + \log\frac{w_\star}{1 - w_\star}\right)\right)\right]\notag\\
    & + w_2\mathbb{E}_z\left[(z + \mu_2 - \mu_1)\sigma\left((\mu_1-\mu_2)^{\top}(z + \mu_2) + \frac{1}{2}(\|\mu_2\|^2_2 - \|\mu_1\|^2_2) + \log\frac{w_1}{w_2}\right)\right]
\end{align}

Eliminate $zg(z)$-like terms with Stein's lemma
\begin{align}
    \nabla_{\mu_1}\mathcal{L} = & w_1 \mathbb{E}_z\left[\mu_1\sigma\left((\mu_1-\mu_2)^{\top}(z + \mu_1) + \frac{1}{2}(\|\mu_2\|^2_2 - \|\mu_1\|^2_2) + \log\frac{w_1}{w_2}\right)\right.\notag\\
    &\left.\left(2 - \sigma\left((\mu_1-\mu_2)^{\top}(z + \mu_1) + \frac{1}{2}(\|\mu_2\|^2_2 - \|\mu_1\|^2_2) + \log\frac{w_1}{w_2}\right)\right)\right.\notag\\
    & + \left.\mu_2\sigma\left((\mu_2 - \mu_1)^{\top}(z + \mu_1) + \frac{1}{2}(\|\mu_1\|^2_2 - \|\mu_2\|^2_2)+ \log\frac{w_2}{w_1}\right)^2 \right.\notag\\
    & + \left. \mu_\star\left(1 - 2\sigma\left(2\mu_\star^{\top}(z + \mu_1) + \log\frac{w_\star}{1 - w_\star}\right)\right)\right]\notag\\ 
    & + w_2\mathbb{E}_z\left[(\mu_2 - \mu_1)\sigma\left((\mu_1-\mu_2)^{\top}(z + \mu_2) + \frac{1}{2}(\|\mu_2\|^2_2 - \|\mu_1\|^2_2) + \log\frac{w_1}{w_2}\right)^2\right]
\end{align}
\section{Numerical experiments settings}
\label{appendix:numerics}
\subsection{Normalizing flows}
\label{sec:app:normalizing}

\paragraph{RealNVPs ---}
One of the ways to construct expressive but invertible functions $f_\theta$ is called RealNVP~\citep{dinh2017density}. The idea is compose coupling layers. These elementary operations partition coordinates of $x$ using a binary mask $b$ of the same dimension and then perform coordinate-wise translation and scaling with functions $t: \mathbb{R}^d\to\mathbb{R}^d$ and $s: \mathbb{R}^d\to\mathbb{R}^d$, respectively. The transformation and its reverse is
\begin{align}
    \label{eq:nf-transform}
    & y = bx + (1 - b)\left[x\odot e^{s(bx)} + t(bx)\right] \\
    & x = by + (1 - b)\left[y - t(by)\right]\odot e^{-s(by)}
\end{align}
We can then stack several layers of such transformations and remain within the class of invertible functions. 

\paragraph{Monte Carlo estimators of gradients ---} Considering the variational family of a normalizing flow with base distribution $p_z$ and map $f_\theta$, the VI objective for a target with density $p$ is the reverse KL:
\begin{align}
    \label{eq:nf-loss}
    \mathcal{L}(\theta) &= D_{\textrm{KL}}(q_\theta||p)= \mathbb{E}_{x\sim q_\theta}\log\frac{q_\theta(x)}{p(x)} \\
    &= \mathbb{E}_{z\sim p_z}\big[\log p_z(z) - \log\left\lvert\nabla_z f_\theta(z)\right\rvert - \log p(f_\theta(z))\big],\notag
\end{align}
where we have applied in the last line the ``reparametrization trick", viewing the VI loss as an expectation over the base density $p_z$ instead of the flow density $q_\theta$. Since the distribution $p_z$ doesn't depend on $\theta$, we can calculate empirical gradients by sampling a batch of $z$ of size $B$ and then taking the gradient of an average of an expression in the square brackets.


\paragraph{Hyperparameters ---}
In the experiments presented in \cref{sec:numerics}, we use RealNVPs with 6 coupling layers. The scaling network $s$ and translation network $t$ are multi-layer perceptrons with 3 hidden layers and hidden dimension $h = 16$.
We fix $w_1 = \sfrac{2}{3}$, $w_2 = \sfrac{1}{3}$, batch size $B = 128$. We perform optimization with Adam at a learning rate of $10^{-3}$. 
Training were considered converged when loss $\mathcal{L}$ as well as the statistics $\mu_{\pm}$ and $w_{\pm}$ were stabilized in the course of 500 iterations, the number of iterations was roughly 1000-4000, depending on the input dimension $d$

\subsection{Mixture of Gaussians}
For the experiments considering the variational family of mixtures of bimodal Gaussians with consider the same batchsize $B=128$ and perform optimization with Adam at a learning rate of $10^{-3}$. Training were considered converged when statistics $m_{1,2}$, $s$, and $w_{1,2}$ were stable in the course of 200 iterations, the number of iterations was roughly 1000.

\end{document}